\documentclass[10pt]{article}

\usepackage{graphicx,color}
\usepackage{amsfonts}
\usepackage{amsmath, amssymb, amsthm}
\usepackage{algorithm}
\usepackage{algorithmic}
\usepackage{subfigure}
\usepackage{url}
\usepackage{epstopdf}
\usepackage{color}

\renewcommand{\baselinestretch}{1.5}

\def \RR {\mathbf{R}}

\newcommand{\argmin}[1]{{\hbox{\text{Arg}$\underset{#1}{\text{min}}\;$}}}

\renewcommand{\min}[1]{{\hbox{$\underset{#1}{\text{Min}}\;$}}}

\newcommand{\casebegin}{\left\{\begin{array}{ll}}
	\newcommand{\caseend}{\end{array}\right.}

\newcommand{\ltriplebar}{\lvert\kern-0.9pt\lvert\kern-0.9pt\lvert}
\newcommand{\rtriplebar}{\rvert\kern-0.9pt\rvert\kern-0.9pt\rvert}

\newcommand{\um}{\underline{m}}

\newcommand{\ux}{\underline{x}}

\newcommand{\uz}{\underline{z}}

\newcommand{\uC}{\underline{C}}

\newcommand{\uS}{\underline{S}}

\newcommand{\uU}{\underline{U}}

\newcommand{\uX}{\underline{X}}
\newcommand{\uY}{\underline{Y}}

\newcommand{\bA}{\textbf{A}}

\newcommand{\bD}{\textbf{D}}
\newcommand{\bE}{\textbf{E}}

\newcommand{\bG}{\textbf{G}}

\newcommand{\bI}{\textbf{I}}

\newcommand{\bP}{\textbf{P}}

\newcommand{\bQ}{\textbf{Q}}
\newcommand{\bR}{\textbf{R}}

\newcommand{\bW}{\textbf{W}}

\newcommand{\myfigJPG}[5]
{\begin{figure}[htbp]
   \begin{center}
     \includegraphics[height=#3in,width=#4in]{#2.jpg}
     \caption{#5}
     \label{fig:#1}
   \end{center}
\end{figure}}

\newcommand{\myfigJPGTwo}[6]
{\begin{figure}[htbp]
   \begin{center}
     \includegraphics[height=#4in,width=#5in]{#2.jpg}\\
     \includegraphics[height=#4in,width=#5in]{#3.jpg}
     \caption{#6}
     \label{fig:#1}
   \end{center}
\end{figure}}

\newcommand{\myfigJPGThree}[7]
{\begin{figure}[htbp]
   \begin{center}
     \includegraphics[height=#5in,width=#6in]{#2.jpg}\\
     \includegraphics[height=#5in,width=#6in]{#3.jpg}\\
     \includegraphics[height=#5in,width=#6in]{#4.jpg}
     \caption{#7}
     \label{fig:#1}
   \end{center}
\end{figure}}

\newcommand{\myfigJPGThreeHor}[7]
{\begin{figure}[htbp]
   \begin{center}
     \includegraphics[height=#5in,width=#6in]{#2.jpg}~
     \includegraphics[height=#5in,width=#6in]{#3.jpg}~
     \includegraphics[height=#5in,width=#6in]{#4.jpg}
     \caption{#7}
     \label{fig:#1}
   \end{center}
\end{figure}}

\newcommand{\myfigJPGFour}[8]
{\begin{figure}[htbp]
   \begin{center}
     \includegraphics[height=#6in,width=#7in]{#2.jpg}\\
     \includegraphics[height=#6in,width=#7in]{#3.jpg}\\
     \includegraphics[height=#6in,width=#7in]{#4.jpg}\\
     \includegraphics[height=#6in,width=#7in]{#5.jpg}
     \caption{#8}
     \label{fig:#1}
   \end{center}
\end{figure}}

\newcommand{\myfigJPGFive}[9]
{\begin{figure}[htbp]
   \begin{center}
     \includegraphics[height=#7in,width=#8in]{#2.jpg}\\
     \includegraphics[height=#7in,width=#8in]{#3.jpg}\\
     \includegraphics[height=#7in,width=#8in]{#4.jpg}\\
     \includegraphics[height=#7in,width=#8in]{#5.jpg}\\
     \includegraphics[height=#7in,width=#8in]{#6.jpg}
     \caption{#9}
     \label{fig:#1}
   \end{center}
\end{figure}}

\begin{document}
\renewcommand{\baselinestretch}{1.2}
\title{Style-Transfer via Texture-Synthesis}

\author{Michael Elad and Peyman Milanfar
\\{\small Google Research}}

\maketitle
\begin{abstract}
Style-transfer is a process of migrating a style from a given image to the content of another, synthesizing a new image which is an artistic mixture of the two. Recent work on this problem adopting Convolutional Neural-networks (CNN) ignited a renewed interest in this field, due to the very impressive results obtained. There exists an alternative path towards handling the style-transfer task, via generalization of texture-synthesis algorithms. This approach has been proposed over the years, but its results are typically less impressive compared to the CNN ones.

In this work we propose a novel style-transfer algorithm that extends the texture-synthesis work of Kwatra et. al. (2005), while aiming to get stylized images that get closer in quality to the CNN ones. We modify Kwatra's algorithm in several key ways in order to achieve the desired transfer, with emphasis on a consistent way for keeping the content intact in selected regions, while producing hallucinated and rich style in others. The results obtained are visually pleasing and diverse, shown to be competitive with the recent CNN style-transfer algorithms. The proposed algorithm is fast and flexible, being able to process any pair of content + style images.

\end{abstract}

\noindent {\bf keywords:} Style-Transfer, Texture-Synthesis, Patch Matching, Segmentation, Convolutional Neural Networks, Tree Nearest-Neighbor, Image Segmentation.

\noindent {\bf EDICS:} ARS-SRV.

\renewcommand{\baselinestretch}{1.5}
% \pagebreak

%%%%%%%%%%%%%%%%%%%%%%%%%%%%%%%%%%%%%%%%%%%%%%%%%%%%%%%%%%%%%%%%%%%%%%%%%%%%%%%%

\section{Introduction}

Style-transfer is a process of migrating a style from one image (the {\em Style-Image}) to another (the {\em Content-Image}). The goal is to synthesize a new image which is an artistic mixture of content and style. In order to illustrate this task, Figure \ref{fig:Examples} presents the style-transfer results obtained by the algorithm proposed in this work, applied on content+style pairs\footnote{As the main message of this work are the the visual results obtained, we bring numerous additional simulation outcomes in an auxiliary part positioned towards the end of this paper.}.

\myfigJPG{Examples}{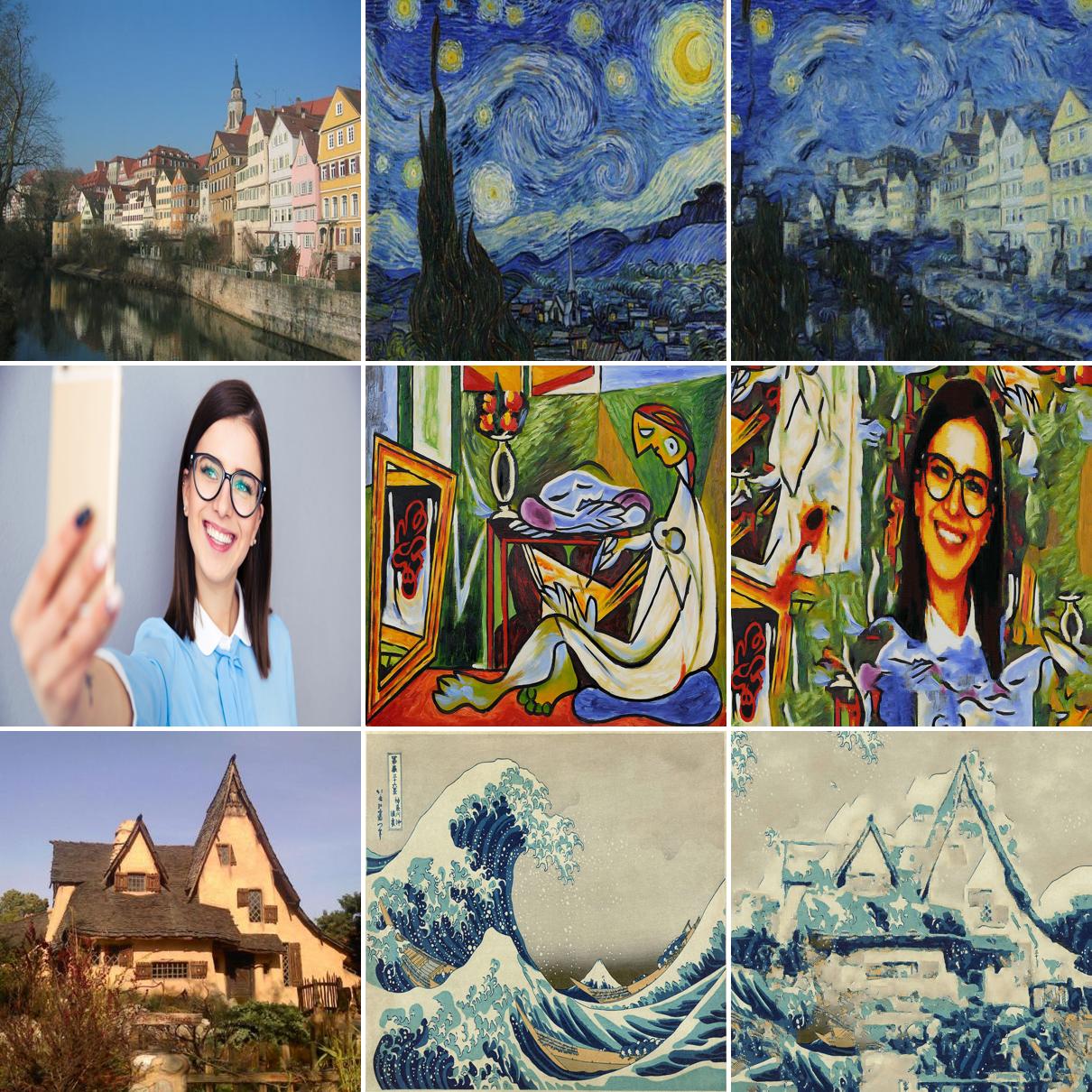}{4.5}{4.5}{{\bf Style-Transfer Examples:} Content image (left), Style image (middle), and the style-transferred results (right) obtained by the proposed algorithm.}

At its core, this transfer task is not well-defined, and as such it could be interpreted and addressed in various ways. More specifically, this fusion is not accompanied by clear answers to questions such as:
\begin{itemize}
\item Which parts of of the content should be preserved (and to what extent) and which discarded or modified?
\item Should the content image be allowed to modify its local contrast as part of the transfer?
\item Should edges and/or other elements in the content image be allowed to shift and if so how?
\item Which color palette should the output adopt?
\item How far should the hallucination be allowed to go in the regions where content is of no importance?
\item Which parts from the style-image qualify as 'style' to be used, and which should be disregarded?
\item Where do we draw the line between 'copying' of style and 'hallucination' of it?
\item Perhaps the most important question of all: What constitutes a successful style-transfer result?
\end{itemize}
\noindent Clearly, these are fundamental issues that govern the outcome to expect from style-transfer methods. Indeed, various algorithms were proposed over the years for handling this problem, each based on a different view of the style-transfer goal. These methods provide either explicitly, or much more often, implicitly, answers to these questions, in their attempt to obtain visually pleasing fusion results.

Broadly speaking, the literature offers two distinct approaches for handling the style-transfer problem: The first is generalization of classic Texture-Synthesis methods, e.g., as practiced in \cite{Efros2001,Frigo}. These algorithms are based on local patch-matching and their aggregation. An  alternative path towards handling style-transfer emerged recently, basing the transfer process on Convolutional Neural Networks (CNN) \cite{CNN1,CNN2,CNN3,CNN4}. Interestingly, these methods as well draw their core intuition from texture-synthesis, but one that is obtained using convolutional neural networks \cite{CNN0}. There are also a few exceptions to these two main activities \cite{AlternativeTrans0,AlternativeTrans1,AlternativeTrans2,AlternativeTrans3,AlternativeTrans4,AlternativeTrans5,AlternativeTrans6,AlternativeTrans7,Survey}. While we do not aim to give here a broad survey of all these existing methods, we shall expand on few milestone contributions in this field in Section \ref{sec:Background}, where we discuss past work, and especially contributions that are relevant to the algorithm proposed in this paper.

Our work addresses the style-transfer problem by returning to the classic texture-synthesis route, leaning on the vast knowledge accumulated in this field, while aiming to get visual results that are on par with recently proposed methods. Indeed, one could consider the style-transfer task as a generalization of texture-synthesis, in which the content image influences the otherwise regular synthesis process. This view has been proposed already in the seminal work by Efros and Freeman (2001) to a limited extent \cite{Efros2001}. More recently, Frigo et. al. (2016) proposed the `Split and Match' approach for achieving the same goal, via an adaptive Quad-Tree division of the content image domain into patches, and quilting matched-pieces taken from the style merged by belief-propagation \cite{Frigo}. The results in this work are far better than those of \cite{Efros2001}, getting closer to those obtained by Gatys et. al. \cite{CNN1}, while being based on a much simpler procedure. Still, the obtained images do not contain strongly hallucinated style elements, as the algorithm forces patches from the content to match those of the style.

In this work we propose a novel style-transfer algorithm that relies on the texture-synthesis work of Kwatra et. al. \cite{Kwatra2005}. This method has been chosen as the foundation for our algorithm due to its elegance, simplicity, and excellent results. Still, we believe that alternative texture-synthesis methods could be adapted for the transfer goal in a similar fashion. We modify Kwatra's algorithm in several key ways in order to achieve the desired style transfer. These include:
\begin{enumerate}
\item An initialization of the algorithm by the content image, augmented by very strong noise, in order to both tie the result to the content in selected areas, while enabling it to depart from it elsewhere.
\item Applying color-transfer from the style to the content within the iterative process, in order to preserve the richness of the style in the final outcome and avoid repetitive patterns, and perhaps the most important of all,
\item Merging the intermediate result with the content image in selected areas, applied in the patch-aggregation step in each iteration. This is based on a segmentation algorithm that defines the importance of content regions.
\end{enumerate}
\noindent Indeed, one interesting positive feature of our solution is that if our algorithm is applied to an empty content-image, it reduces gracefully to be Kwatra's classic texture synthesis algorithm, applied on the style image. Figure \ref{fig:NoContent} presents such hallucination results for several style images.

\myfigJPGThree{NoContent}{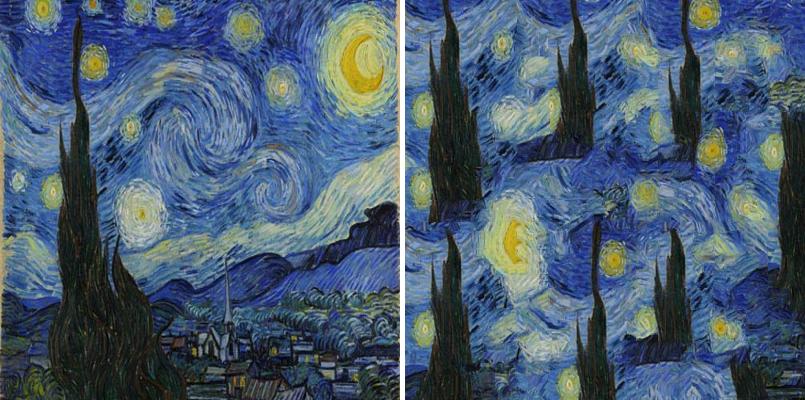}{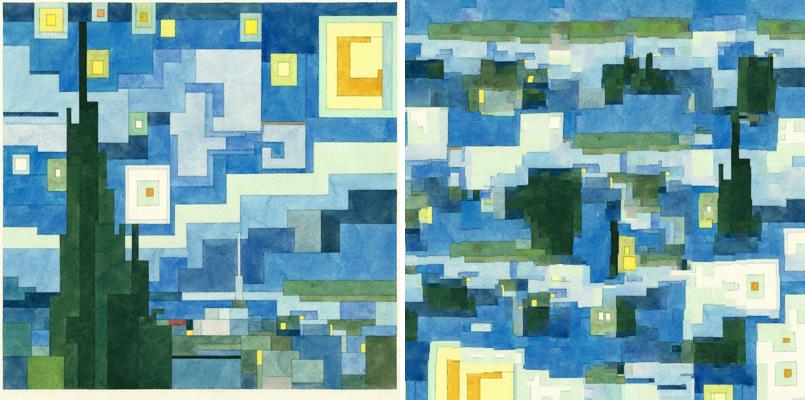}
{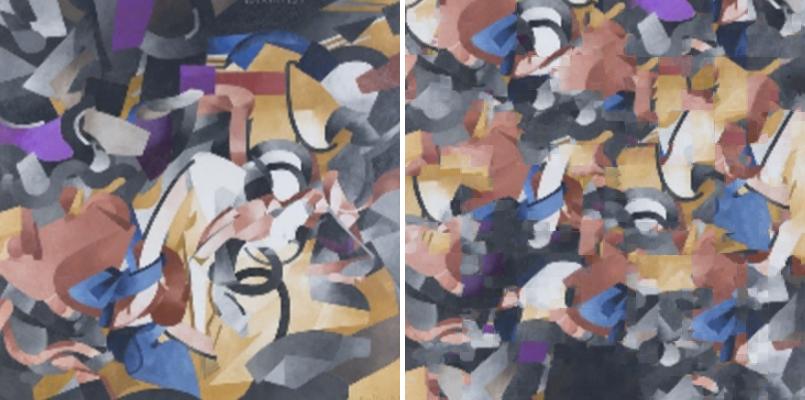}{1.6}{3.2}{{\bf No-Content Style-Transfer:} Style image (left), and its hallucination result (right) by Style-transfer applied with an empty content image.}

Addressing the questions posed above, our algorithm's answers are far more explicit than those given by CNN methods, and these answers are the following:
\begin{itemize}
\item Content to be preserved? We answer this by proposing the use of a segmentation algorithm, deployed on the content image as a pre-process, and defining clearly the regions to be preserved and the extent of this preservation. Local contrast modifications are allowed as part of a pre-process of the input content image.
\item Shift of edges and/or other elements? Our answer is mostly negative - edges and other objects in the segmented areas remain in their original locations in the content image, but may deviate slightly due to the patch-matching applied.
\item Color palette? While our method has the flexibility of transferring the color from the content to the style, or adopt any other palette of choice, the results we shall present in this paper correspond to histogram matching that transfers the palette from the style to the content only. The reason for this choice is the tendency of style images to be much richer in their color expression, thus offering a more stable palette to rely upon.
\item How far can the hallucination go? As opposed to other classic texture-synthesis based methods (and even the recent one -- \cite{Frigo}), we allow the created image to depart from the content-image, and create random combinations of elements taken from the style, as seen in Figure \ref{fig:Examples}. This is one of the prime sources for the high-quality and artistic results obtained with our method.
\item Parts in the style-image qualifying as 'style'? We consider every detail in the style image as candidate for being used in the fusion outcome. The chosen pieces are eventually decided upon based on the randomized matchings done, and influenced by the regions in the content that are allowed to drift. This answer should not be taken lightly, as, for example, if the style-image contains faces, they are likely to appear in the stylized result, and we will consider this as a successful outcome. If this is an undesired effect, it can be easily controlled by an additional mask that marks the permitted regions to be used from the style. While we do not provide examples for such masking, it is a simple extension to the work reported here.
\item Copying versus hallucination? Being a texture-synthesis based algorithm, our method essentially merges patches from the style-image. This might be substantially different from CNN-based methods that are likely to synthesize unseen structures (a questionable property). Therefore, elements of the style are copied in our method. Nevertheless, measures for avoiding copy of large pieces are taken, by pushing the image to keep its color diversity, and by working with varying scales and varying patch sizes.
\item Successful results? While the proposed algorithm is robust and consistent, it does not mean that each produced result can be considered as a successful one. The quality of the outcome depends primarily on the success of the segmentation algorithm to keep the important content from being overridden. Other factors that govern the output quality are the match between the content image and the style in terms of the palette imposed, and the match between the content details and the scale of objects found in the style image. We shall elaborate on each of these factors in the results section, accompanying these explanations with failure examples.
\end{itemize}

\noindent The results we obtain with the proposed algorithm are competitive with the recent style-transfer algorithms, CNN \cite{CNN1,CNN2,CNN3,CNN4} or texture-synthesis based \cite{Frigo}. The created images are visually pleasing, containing rich and diverse hallucinated parts brought from the style, while keeping the essence of the content intact. In terms of the computational complexity, the most demanding part of the algorithm is the patch-matching that takes place in the very last full-resolution iterations. The run-time for the whole process in a direct Matlab (no Mex) implementation stands on $25-50$ seconds for an image of size $400 \times 400$ pixels.

This paper is organized as follows: In Section \ref{sec:Background} we describe past work related to the task of style-transfer, aiming to provide background to the method we present in this paper. Section \ref{sec:Algorithm} outlines all the ingredients of our proposed algorithm, tying them carefully to Kwatra's work \cite{Kwatra2005}. Section \ref{sec:Results} brings series of results, demonstrating the various aspects of the proposed algorithm and showing its success and failure results. In this Section we also discuss several key features of the proposed method in light of the results shown. In Section \ref{sec:Conclusions} we conclude this work and highlight possible extensions. An appendix with more examples is given at the end of the paper.

%%%%%%%%%%%%%%%%%%%%%%%%%%%%%%%%%%%%%%%%%%%%%%%%%%%%%%%%%%%%%%%%%%%%%%%%%%%%%%%%

\section{Past Work on Style-Transfer}\label{sec:Background}

One of the earliest works related to our topic of style-transfer is reported in \cite{Efros2001}. This paper by Efros and Freeman belongs to the branch of texture-synthesis based methods -- algorithms that aim to synthesize textured images based on a given texture example (e.g. see \cite{TexSyn1,TexSyn2,TexSyn3,TexSyn4}). This paper suggested a patch-quilting procedure for texture synthesis, and then shown how to extend it to target a closely related goal of texture-transfer (rather than style-transfer). Their view of the style-transfer task was more conservative, assuming that a general shades of the given content image are to be reproduced by tiling patches from a given texture image. Merging the patches over their overlaps was proposed to be done by finding optimal seam-curves between adjacent patches which exhibit minimum variation. Later work by Freeman in a different context (example-based super-resolution) \cite{FreemanBP1,FreemanBP2} replaced this patch merging procedure with a better one based on a Belief-Propagation (BP) approach. BP serves as a combinatorial approximate solver for an energy minimization procedure, handling the task of seeking the best patch assignments among few neighbors per each location so as to get the smoothest outcome.

These works have been followed by many, most of which studied the problem of texture-synthesis, aiming to generate a pseudo-random texture image from a given texture example with a high visual quality outcome. One of these follow-up works is the paper by Kwatra et. al. \cite{Kwatra2005}, which proposed a very effective texture-synthesis approach, relying on the same core principles as in \cite{Efros2001} of patch matching and their fusion. Kwatra et. al. adopted a global optimization point of view, of seeking an overall synthesized image that would give the minimal accumulated local distance to patches extracted from the example texture. This task has been broken into two stages, in the spirit of the EM (or K-Means clustering) algorithm \cite{EM}, in the first stage freezing the current global image and seeking the best patch-matches, and in the second freezing the found matches and seeking to update the global image aggregating all these chosen patches. Their proposed algorithm has been shown to lead to high quality texture-synthesis when applied sequentially on several patch sizes from large to small, iterating the EM method several times per each. Further improvement to this method is their suggestion to operate on a Gaussian pyramid of the destination image. As we shall see in section \ref{sec:Algorithm}, we shall adopt all these ideas and extend them to develop our style-transfer method.

A parallel activity that addressed synthesis of stylized images is the one based on {\em analogies}, e.g., the work reported in \cite{Analogy1,Analogy2,Analogy3,PatchTable}. The problem statement is somewhat different from the one discussed here, as it relies on the availability of an input image and its stylized result. Based on such an example pair, given a new content image we aim to apply the transfer in analogy to the example-pair given. We shall not discuss this branch of activity further in this work, as it deviates from our problem.

We skip more than a decade to describe a very recent and impressive paper by Frigo et. al. \cite{Frigo} that suggests a style-transfer algorithm via texture-synthesis. Their approach, to a large extent, can be considered as a direct followup of the work reported in \cite{Efros2001,FreemanBP1,FreemanBP2}, adapted to include the influence of the content image. Thus, whereas these papers used fixed-size patches and belief-propagation to merge them properly, Frigo et. al. offer to divide the image domain adaptively into squares, driving this adaptive quad-tree division by the distance to the nearest neighbor in the style image and the inner variance of the content patch. The results obtained by this method are far better than those appearing in \cite{Efros2001}, but still lack hallucination ability, as each patch must be close enough to the content patch it covers. In that respect, this work's results tend to be more of a texture- than style-transfer.

In 2015 Gatys et. al. proposed a brilliant and very different style-transfer method that was shown to lead to very impressive results \cite{CNN1}. Their approach to the problem was the first to adopt a pre-trained CNN \cite{VGG}, as the means for extracting features from both the style and the content images. Interestingly, their work was based on earlier effort to use CNN for handling the texture-synthesis problem \cite{CNN0}, and so we see the same pattern of building a texture-synthesis algorithm, and then generalizing it to obtain style-transfer.

Gatys et. al. posed the style-transfer problem as an energy-minimization task, seeking an image close to the content one, while also providing correlation-map that is close to that obtained by the style image -- all measured in the VGG feature domain \cite{VGG}. This method essentially reverses the CNN, seeking an input image based on the features desired, and as such, leading to quite a demanding numerical optimization. Note that while the objective function in Gatys' work is very clearly defined, it serves only as a proxy to answer the questions posed above about the nature of the fusion to apply. Results of this method were shown to be very daring, going well beyond simply texturizing the content image (see Figure \ref{fig:Examples}), and allowing the creation of images that could be perceived as high-quality art. Naturally, this ignited a renewed interest in the problem of style-transfer, and especially in the use of CNN for this purpose. The work reported in \cite{CNN2,CNN3,CNN4} present various improvements, mainly in speed, over Gatys' method, by training a feed-forward network to achieve the overall goal, whiletraining it using Gatys' penalty function. We will not expand on this further. While we do not adopt any of the features that characterize these algorithms in our proposed method, we are strongly influenced and inspired by the kind of results these methods obtain, and we shall aim to get close to the flavor of those results while keeping the algorithm a classic (patch-matching) texture-synthesis oriented one.

%%%%%%%%%%%%%%%%%%%%%%%%%%%%%%%%%%%%%%%%%%%%%%%%%%%%%%%%%%%%%%%%%%%%%%%%%%%%%%%%
%%%%%%%%%%%%%%%%%%%%%%%%%%%%%%%%%%%%%%%%%%%%%%%%%%%%%%%%%%%%%%%%%%%%%%%%%%%%%%%%

\section{The proposed Algorithm}\label{sec:Algorithm}

In this section we describe in details the proposed style-transfer algorithm. As explained earlier, many of its ingredients are borrowed directly from Kwatra et. al. \cite{Kwatra2005}, and thus we shall clearly define the changes that brought us to handle the transfer task.

%%%%%%%%%%%%%%%%%%%%%%%%%%%%%%%%%%%%%%%%%%%%%%%%%%%%%%%%%%%%%%%%%%%%%%%%%%%%%%%%

\subsection{Energy Minimization Point-of-View}

We start by defining the core objective of our algorithm: We are given a content image\footnote{All the computations in our algorithm are performed in the RGB domain, and thus the factor 3 in the definition of the content and style images.}  $\uC\in \RR^{3N_c}$, and a style image $\uS\in \RR^{3N_s}$. These two images are accompanied by a segmentation mask $\bW\in [0,\infty)^{N_c}$ that marks the importance of pixels in the content image, in terms of its parts to be preserved. More on this mask will be brought later in this Section. Our goal is the creation of the image\footnote{Eventually, $\uX$ should be an image of the same size as $\uC$, but we define a series of optimization tasks that operate in several scales of the problem, and thus $\uX$'s size varies across the algorithm stages.} $\uX$ that would minimize the following series of energy functionals:
\begin{eqnarray}
E_{L,n} \left\{X\right\}\label{eq:objective1}
= \frac{1}{c}  \sum_{(i,j) \in \Omega_{L,n}}  \min{(k,l)}
\left\|  \bR_{ij}^{n} \uX - \bQ_{kl}^{n} \bD_L^S \uS \right\|_2^r + \| \bD_L^C \uC -\uX \|_\bW^2  + \lambda \rho\{\uX\}.
\end{eqnarray}
The first term is very similar to the formulation of the problem as posed in Kwatra's work \cite{Kwatra2005}, although it may seem little bit different due to our notations. The operators $\bD_L^C$ and $\bD_L^S$ stand for down-scaling\footnote{Since these two image may be of different sizes, different matrix operators are attributed to each.} of $\uX$ (and $\uC$) and $\uS$, for operating over a Gaussian pyramid of our images. Assume for now that we operate in the native resolution of the images, using $\bD_1^C =\bI$ and $\bD_1^S=\bI$, and thus our goal is
\begin{eqnarray}
E \left\{X\right\}\label{eq:objective2}
= \frac{1}{c}  \sum_{(i,j) \in \Omega}  \min{(k,l)}
\left\|  \bR_{ij}^{n} \uX - \bQ_{kl}^{n} \uS \right\|_2^r + \| \uX- \uC \|_\bW^2  + \lambda \rho\{\uX\}.
\end{eqnarray}
The operator $\bR_{ij}^{n}$ denotes the extraction of a patch of size $n$ from location $(i,j)$ in the image $\uX$. This term expresses our desire that the image $\uX$ should be such that every patch of size $n$ extracted from it from location $(i,j)$ would be close to a patch of the same size extracted from $\uS$. The operator $\bQ_{kl}^{n}$ represents the patch extraction from the style image, and within this minimization procedure, both $\uX$ and all the assignments $(k,l)$ (for patches of size $n$) are considered as unknowns. The coefficient $1/c$ normalizes the first term by taking into account the level of overlap between the aggregated patches in $\uX$.

The summation over $(i,j)$ is done over the domain $\Omega$. While this domain could consider fully overlapped patches and cover each and every pixel in the support of the image $\uX$, we choose to force these matchings over a decimated grid of points, in order to reduce computations. This means that we skip a constant number of rows/columns from one patch-match to the other. Also, in the same spirit as in \cite{Kwatra2005} of seeking a robust patch aggregation over the overlaps, the local patch errors are put to the power $r$ ($=0.8$) instead of the usual $L_2$ squared term.

We now bring back the two hyper parameters governing the energy functional -- $n$ and $L$. These correspond to our desire to force the matching of $\uX$ to patches from the style image using several patch-sizes $n$, and over several scales ($L=1,2,~\ldots~,L_{max}$). The patch sizes we use are $n_1,~ n_2,~\ldots~,~ n_m$, and this guarantees that several element-sizes of the style features are adopted in the style-transfer. Furthermore, the entire penalty is enforced over several scales of the content and style images, in order to get better minimization. Each choice of these two parameters yields a different energy function, and our overall algorithm will handle this chain of functionals sequentially in order to obtain the final outcome.

As for the other parts of the expression in Equation (\ref{eq:objective1}), the second term enforces the content parts marked by the segmentation map of the final image, in the proper scale of the pyramid. The third term is the ($-$) log of the prior, forcing $\uX$ to obey general image statistics, essentially driving $\uX$ (in all the resolution layers of its pyramid) to be spatially smooth.

Our goal is to find the image minimizing the above energy functional for the smallest patch-size $n_m$ and the maximal resolution layer $L=1$. In order to better direct the solver and avoid local minimum, our approach is iterative and sequential, minimizing a chain of intermediate energy functionals that sweep through the scales and the patch-sizes, generating a path of refined solutions this way. For simplicity of the presentation, we shall concentrate at the moment on a subproblem in which only one patch-size is used, and the penalty consider only the native resolution of the handled images (i.e. no pyramid decomposition). Later on we shall bring these two effects to the discussion for the completeness of the algorithm description.

%%%%%%%%%%%%%%%%%%%%%%%%%%%%%%%%%%%%%%%%%%%%%%%%%%%%%%%%%%%%%%%%%%%%%%%%%%%%%%%%

\subsection{The EM Optimization Structure}

With the omission of the patch-sizes and the pyramid decomposition, our target optimization problem is therefore,
\begin{eqnarray}
\min{X} \frac{1}{c} \sum_{(i,j) \in \Omega} \min{k,l} \|  \bR_{ij}^{n} \uX - \bQ_{kl}^{n} \uS \|_2^r + \| \uX- \uC \|_{\bW}^2  + \lambda \rho\{ \uX\}.
\end{eqnarray}
The approach we shall take in handling this problem is block-coordinate-descent, in which we sequentially freeze some of the unknowns and update the others, and repeat this strategy for several iterations. Kwatra et. al. suggested the very same strategy, explained to be an Expectation-Maximization treatment to the problem at hand \cite{Kwatra2005}.

First, assume that $\uX$ is fixed and known, and our goal is to find the best matched-patches from the style image. In this case the problem decomposes into a set of separable optimization tasks of seeking for each patch $\bR_{ij}^{n} \uX$  its closest neighbor in the style image. These problems read
\begin{eqnarray}
\{k^*,l^*\}=\min{(k,l)} \left\|  \bR_{ij}^{n} \uX - \bQ_{kl}^{n} \uS \right\|_2,
\end{eqnarray}
and the patch matched to $\bR_{ij}^{n} \uX$ would be $\uz_{ij} = \bQ_{k^*,l^*}^{n} \uS$. The distance measure to be used is $L_2$, and we shall discuss fast ways to implement this search later in this Section.

Having found the best matches to the style image, those are kept fixed, and our goal now is to update $\uX$ by solving
\begin{eqnarray}
\min{\uX} \frac{1}{c} \sum_{(i,j) \in \Omega} \left\|  \bR_{ij}^{n} \uX - \uz_{ij} \right\|_2^r + \left\| \uX- \uC \right\|_{\bW}^2  + \lambda \rho\{ \uX\}.
\end{eqnarray}
We handle this task by the Plug-and-Play approach \cite{PPP}. First, we modify the format of the problem to be
\begin{eqnarray}
\min{\uX,\uY} \frac{1}{c}  \sum_{(i,j) \in \Omega} \left\|  \bR_{ij}^{n} \uX - \uz_{ij} \right\|_2^r + \left\| \uX- \uC \right\|_{\bW}^2  + \lambda \rho\{ \uY\} ~~~s.t. ~~~\uY=\uX.
\end{eqnarray}
Then we use the Augmented-Lagrangian algorithm to handle the constraint, resulting with
\begin{eqnarray}
\min{\uX,\uY} \frac{1}{c}  \sum_{(i,j) \in \Omega} \left\|  \bR_{ij}^{n} \uX - \uz_{ij} \right\|_2^r + \left\| \uX- \uC \right\|_{\bW}^2  + \lambda \rho\{ \uY\}+ \mu \left\| \uX-\uY+\uU \right\|_2^2,
\end{eqnarray}
where $\uU$ serves as the Lagrange multiplier vector for the set of constraints. ADMM solution of the above is done by updating $\uX$, $\uY$, and $\uU$ iteratively using the block-coordinate descent \cite{ADMM}, resulting with the following steps:
\begin{itemize}
\item {\bf Step 1:} Update of $\uX$ while keeping $\uY$ and $\uU$ as known,
\begin{eqnarray}
\min{\uX} \frac{1}{c}  \sum_{(i,j) \in \Omega} \left\|  \bR_{ij}^{n} \uX - \uz_{ij} \right\|_2^r + \left\| \uX- \uC \right\|_{\bW}^2  + \mu \left\| \uX- \uY+ \uU \right\|_2^2.
\end{eqnarray}
This problem is handled by the Iterative Reweighed Least Squares (IRLS) \cite{IRLS}, in which we  modify the $r$ power of the $L_2$ into $2$ by adding "pseudo-weights", and then solving the obtained quadratic expression as a simple patch-aggregation procedure. More specifically, in the $k^{th}$ iteration we have the temporary solution ${\hat \uX}_k$ and we compute the weights
\begin{eqnarray}
w_{ij}=\left\|  \bR_{ij}^{n} {\hat \uX}_k - \uz_{ij} \right\|_2^{r-2},
\end{eqnarray}
with which we redefine our task as
\begin{eqnarray}
\min{\uX} \frac{1}{c} \sum_{(i,j) \in \Omega} w_{ij} \left\|  \bR_{ij}^{n} \uX - \uz_{ij} \right\|_2^2 + \left\| \uX- \uC \right\|_{\bW}^2  + \mu \left\| \uX-\uY+\uU \right\|_2^2.
\end{eqnarray}
This has a closed form solution,
\begin{eqnarray}
{\hat \uX}_{k+1} & = & \left[ \frac{1}{c} \sum_{(i,j) \in \Omega} w_{ij} (\bR_{ij}^{n})^T (\bR_{ij}^n) + \bW + \mu \bI \right]^{-1} \\
& & \nonumber ~~~~\cdot\left[  \frac{1}{c} \sum_{(i,j) \in \Omega} w_{ij}  (\bR_{ij}^{n})^T  \uz_{ij} + \bW \uC + \mu (\uY-\uU) \right].
\end{eqnarray}
Putting aside the role of the Lagrange multiplier (assigning $\mu=0$), this expression suggests to aggregate all the patches $\uz_{ij}$ into their locations in the overall image by weighted averaging governed by $w_{ij}$, and then applying a weighted average of this outcome with the content image by selectively choosing the regions highlighted by $\bW$. This whole iterative process should be applied several times ($10$ iterations in our experiments) to obtain the final solution for $\uX$.

\item {\bf Step 2:} Update of $\uY$ while freezing $\uX$ and $\uU$ by
\begin{eqnarray}
\min{\uY} \lambda \rho\{\uY\}+ \mu \|\uX- \uY+ \uU \|_2^2,
\end{eqnarray}
which is simply a denoising procedure applied on the image $\uX+\uU$ with the assumption that the noise level is $\lambda/\mu$. Notice that so far we refrained from defining the choice of $\rho\{\uX\}$, and now the reason becomes clear. The choice of the denoising algorithm is our way of choosing which prior to be used. This way, we may invoke highly effective denoising algorithms without explicitly defining $\rho$. This is the essence of the Plug-and-Play Prior method as proposed in \cite{PPP}. In this work we chose to apply the Domain-Transform filtering \cite{DomainTrans}, a fast approximation algorithm for the bilateral filter.

\item {\bf Step 3:} Update of $\uU$ should be done by the Augmented Lagrangian approach \cite{ADMM}, of $\uU=\uU+\uX-\uY$.
\end{itemize}

\noindent In the special case in which the prior is removed, our problem becomes simpler, and its solution can be approximated by IRLS directly. A further shortcut can be proposed in this case, in which the overall optimization we treat,
\begin{eqnarray}\label{eq:IRLStaskALL}
\min{\uX} \frac{1}{c} \sum_{(i,j) \in \Omega}  \left\|  \bR_{ij}^{n} \uX - \uz_{ij} \right\|_2^r + \left\| \uX- \uC \right\|_{\bW}^2
\end{eqnarray}
could be broken into two, much simpler, parts. In the first of these we minimize the first term only, resulting with the temporary result ${\tilde \uX}$
\begin{eqnarray}\label{eq:IRLStask}
{\tilde \uX}=\argmin{\uX}  \sum_{(i,j) \in \Omega} \left\|  \bR_{ij}^{n} \uX - \uz_{ij} \right\|_2^r.
\end{eqnarray}
This can be done using the very same IRLS as used by Kwatra \cite{Kwatra2005}. Then the solution to the overall problem can be approximated as
\begin{eqnarray}\label{eq:MERGEtask}
{\hat \uX} =( \bW+\bI)^{-1} ( {\tilde \uX} + \bW \uC ).
\end{eqnarray}
Note that this approach is perfectly exact when $r=2$, since in this case we have that $\frac{1}{c} \sum_{(i,j) \in \Omega} (\bR_{ij}^{n})^T (\bR_{ij}^n) = \bI$.

Still in the spirit of simplifying things, we can now bring back the prior and deploy it as a denoising post-process stage on the resulting image ${\hat \uX}$. Indeed, all the results we shall present later on refer to these numerical shortcuts.

%%%%%%%%%%%%%%%%%%%%%%%%%%%%%%%%%%%%%%%%%%%%%%%%%%%%%%%%%%%%%%%%%%%%%%%%%%%%%%%%

\subsection{Multi-Scale and Multi-Patch Sizes}

All the above process should be applied to varying patch-sizes and to several resolution scales of the images involved ($\uX$, $\uC$, and $\uS$). One could imagine merging all these energy functionals together into one holistic term to be minimized, but this is not the path taken here. We deploy a sweep over the patch sizes and over the scales sequentially. This has the spirit of the stochastic gradient descent approach, where each patch-size and resolution level contributes its influence to the final outcome separately, thus leading to a better steady-state final result, with better chances of avoiding local minima.

The proposed sweep over the patch sizes and the scales is done only for one round, starting from the coarsest resolution down to the native one, and for each such resolution sweeping through patch sizes from the largest to the smallest. For each $L$ (resolution level) and $n$ (patch-size) we apply a fixed number of outer iterations to update $\uX$ and the patch-assignments $\{(k,l)\}$, and within each of these we apply $10$ inner iterations to solve the IRLS problem. Each sub-optimization problem is initialized with the output of the preceding optimization, with an option to lead to randomized overall results by adding strong noise to the temporary solution at the beginning of the processs in each resolution later. The very first of all these optimization steps is initialized by the content image with very strong additive Gaussian noise ($\sigma=50$), so as to enable the algorithm to match patches daringly.

This process serves our overall goal of getting a fast end-to-end style-transfer process that gradually refines the result. One refinement effect is in terms of the spatial resolution of the resulting image, as we go from the coarsest to the finest resolution layer. A second refinement effect is obtained by starting with big patches ($33 \times 33$) and moving gradually to smaller and smaller ones ($21\times 21$, $13\times 13$, $9\times 9$, and possibly going down to $5 \times 5$), which gives the ability to the algorithm to adopt large elements from the style image, while refining and modifying them locally. The third and perhaps the most important refinement effect corresponds to the influence of the content image. The details of the content are consistently pushed into the temporary result, weighted by $\bW$, and influencing the hallucination results obtained, delicately in regions where $\bW$ is low, and more pronounced in regions where $\bW$ is high. This way, important regions in the result never depart too far from the content image, while less important regions are allowed to drift and hallucinate.

%%%%%%%%%%%%%%%%%%%%%%%%%%%%%%%%%%%%%%%%%%%%%%%%%%%%%%%%%%%%%%%%%%%%%%%%%%%%%%%%

\subsection{Segmentation and its Role}

While the computation of the segmentation mask $\bW$ could be considered as outside the scope of this work, we do mention briefly several options we experimented with. We should note that each of these methods has its own strengths and weaknesses, and there is no clear choice between them for our purposes. The tested segmentation methods are the following:
\begin{itemize}
\item {\bf Edge-based segmentation:} Per each pixel in $\uC$ we may accumulate all the local gradient vectors as a matrix $\bG$ of size $n^2\times 2$, where the accumulation neighborhood is $n\times n$ pixels. Computing the two singular values of this matrix (there is a closed form expression for these in this case) provides information about the local coherence and contrast (see \cite{RAISR} for more details). Thresholding both these measures, we can obtain a map of the consistent and strong edges in $\bG$. A region filling over this result leads to one segmentation option we have experimented with. We shall refer to this as {\em Edge-Based Segmentation}.

\item {\bf Affinity-based segmentation:} For the image $\uC$ (that has $N_c$ pixels) we can build an $N_c \times N_c$ affinity matrix $\bA$ that corresponds to the interrelations between its pixels. We consider for example the affinity matrix obtained for the bilateral filter \cite{DomainTrans,Hossein}, built by assessing distances between the pixels in both spatial and radiometric domains. Give this matrix, its second eigenvector serves as a candidate segmentation of the image. While all this process may sound computationally daunting, numerical shortcuts based on very sparse sampling of the rows of $\bA$ and the Nystrom method, make all this process fast and effective. We shall not expand on this approach here, as it is not included in the reported experiments.

\item {\bf Face-detection based segmentation:} Most of the images taken by cell-phones and which would be candidates for style-transfer contain faces. One would even go to the extreme of claiming that most of these are {\em selfies}. The above two segmentations are entirely oblivious to faces, and this means that we may ruin identifiability of the people photographed if not segmented well. For such cases, our experiments included yet another segmentation method that starts by detecting faces, and then searches for the bodies to connect to these faces, finalizing the segmentation by a {\em grabcut} algorithm \cite{GrabCut}. We shall refer to this as {\em Face-Based Segmentation}.
\end{itemize}

\noindent Could we do without segmentation altogether? The answer is not conclusive. In the results section we shall present such results in which $\bW$ is set to be a constant image of value $0<\alpha<\infty$. When $\alpha=0$, we get texture-synthesis, hallucinating an image built from the style image alone, without any influence of the content image. When $\alpha\rightarrow \infty$, we force the content so strongly so as to prevent the algorithm to achieve any transfer of the style. For intermediate values we do get texturization of the image, but our experiments indicate that often times, this process does not end with visually pleasing results.

One may question how CNN-based style-transfer algorithms can operate without an explicit segmentation. We tend to believe that the answer is that somehow this segmentation exists within the VGG network implicitly. This highlights two interesting and critical differences between our solution and CNN based ones:
\begin{enumerate}
\item On the positive side, our algorithm is built of clear and simple building blocks, and as such, the influence of its ingredients (e.g., the chosen patch-sizes, segmentation and its effect, and practically every other piece of the algorithm) on the final outcome is rather clear. This helps in adjusting the algorithm's parameters and in providing the user with meaningful knobs to control the results obtained.
\item On the negative side, CNN based style-transfer methods permit themselves to go into the important content parts and modify them. Most often, this is done delicately, leading to pleasing results. Our current solution is slightly different as it is likely to less influence these regions, providing only a delicate texturizing effect. We consider this as a shortcoming of our approach and we intend to invest more work in addressing it in our future work.
\end{enumerate}

%%%%%%%%%%%%%%%%%%%%%%%%%%%%%%%%%%%%%%%%%%%%%%%%%%%%%%%%%%%%%%%%%%%%%%%%%%%%%%%%

\subsection{Color Transfer}

The above described algorithm operates fully in the RGB domain. One may envision modifying it to work only on the luma channel, while modifying the chroma channels independently, but we leave this for future work.

As already mentioned in the Introduction, the resulting image palette could be controlled to be anything desired. In particular, it could be adopted from the style, kept faithful to the content image, or be anything else chosen. The key is to pre-process the incoming images $\uC$ and $\uS$, bringing them both to common grounds of sharing the same palette before the optimization process starts. While we experimented with several such options, we found that in most cases the palette of $\uS$ is far richer than that of $\uC$, and thus aligning with it tends to lead to better visual results.

Palette matching is critical not just as a pre-process stage, but also within the iterative refinement algorithm. The reason is the desire to force the algorithm to preserve the diversity offered by the style image, avoiding a partial use of its wealth. Indeed, each round (optimization of $E_{L,n}$ for a specific resolution layer and patch-size) is followed by rematching the result to the destination palette. As an example, in the popular starry-night style image, there are yellowish stars of different sizes. Without the re-projection onto the palette, a solution may avoid the use of the yellow parts altogether, giving a less daring result.

The actual palette matching can be done by a direct histogram matching, by a simpler parametric method that aligns the moments of the colors span, and other techniques \cite{Color1,Color2,Color3,Color4}. All the results shown in this work were obtained with the histogram matching supported by Matlab (\textsf{imhistmatch}). In some cases in which the style image is not rich enough or tends to overemphasize a specific unnatural color, this may inflict on the results, rendering them unpleasant. Another possible problem takes place when the style image has a large dark region. In such cases we essentially force the same relative darkness on the output images, often times resulting with unpleasant results.

The above comments suggest that prior to activating the style-transfer algorithm, the chosen style images should be carefully picked and pre-processed, so as to increase chances of success of the transfer task. Still in the same spirit to this line of thinking, our algorithm considers all the elements within the style image as {\em style}, without an ability to distinguish between artistic styled patches and other objects. As such, if, for example, the style image contains faces or other distinguishable features, they are very likely to be used in the style-transfer, appearing in random locations and in different contexts. As mentioned in the Introduction, this can be easily managed by masking some regions of the style image.

%%%%%%%%%%%%%%%%%%%%%%%%%%%%%%%%%%%%%%%%%%%%%%%%%%%%%%%%%%%%%%%%%%%%%%%%%%%%%%%%

\subsection{Fast Nearest Neighbor (NN) Used}

The most computationally intensive part of the proposed algorithm is the patch matching, solving problems of the form
\begin{eqnarray}\label{eq:NNtask}
(k*,l*)=\min{(k,l)} \|  \bR_{ij}^{n} \uX - \bQ_{kl}^{n} \uS \|_2.
\end{eqnarray}
Posed differently, the patches of the style image can be pre-organized for each resolution layer and each patch size, into a matrix $\bP_{L,n}$ of size $n\times M_{L,n}$, where $M_{L,n}$ is the number of possible patches. Denoting $\ux_{ij}=\bR_{ij}^{n} \uX$, our goal is to find the column in $\bP_{L,n}$ that is closest to this vector. Many methods exist for handling this problem efficiently (e.g., \cite{NN1,NN2,NN3,NN4,NN5,NN6,NN7,NN8,PatchTable}), and it is beyond the scope of this work to explore all these options. Our approach to the problem is to reduce dimensions in both axes of $\bP$, reducing the vectors dimensions, $n$, and reducing the need to linearly compare $\ux_{ij}$ with each of the $M_{L,n}$ columns.

The data dimension $n$ is reduced by PCA, removing the mean $\um _P$ vector from all the columns of\footnote{We omit hereafter the dependency on $L$ and $n$ to keep the explanation simple.} $\bP$ and finding the leading $k(\ll n)$ eigenvectors spanning the obtained residuals, forming the rows of the projection matrix $\bE_P$. The destination dimension is chosen so as to preserve $95\%$ of the energy of the original data, resulting with a projection matrix of size of size $k\times n$. In the pre-process stage, all the style patches are projected by computing  ${\tilde \bP} = \bE_P(\bP-\um _P \underline{1}^T)$, and this should be done for all patch sizes and all resolution layers.

Given a vector $\ux$ for which we seek its NN, we first condition it to the PCA domain of the data by ${\tilde \ux} = \bE_P(\ux-\um_P)$, and then search for NN within ${\tilde \bP}$ in this reduced dimensionality. Once the NN is found, the true patch from $\bP$ is used as $\uz_{ij}$ for our further processing.

As for the need to reduce the number of examples, we follow the recommendation made by Kwatra et. al. \cite{Kwatra2005}, of approximating this search by constructing a clustering tree. In our implementation, our tree allows for overlaps between the subsets in order to avoid inaccuracies (i.e. $50\%$ of the examples in each stage of the tree clustering are allowed to participate in more than one cluster).

%%%%%%%%%%%%%%%%%%%%%%%%%%%%%%%%%%%%%%%%%%%%%%%%%%%%%%%%%%%%%%%%%%%%%%%%%%%%%%%%

\subsection{An Overall Description of the Proposed Algorithm}

In Figure \ref{fig:algo} we outline the complete structure of the proposed algorithm, with all its ingredients. As can be seen, we start from the coarsest resolution and slowly work our way to the refined layers. The image $\uX$ is being updated and scaled-up as we pass from one layer to the next. The output of the algorithm is the image $\uX$, but without the last stage of the weighted average of $\uC$, so as to leave the effect of delicate textured regions also in the content parts.

\begin{figure}[hbtp]
\begin{center}
\begin{tabular}{|c|}
\hline
\begin{minipage}[b]{0.9\linewidth}
    \renewcommand{\baselinestretch}{1}
    \small \vspace{0.1in}

    \begin{description}
    \item {\bf Objective:} Create the style-transfer image, $\uX$.

    \item {\bf Input:} Use the following ingredients and parameters:
    \begin{itemize}
            \item $\uC$ and $\uS$ - Content and Style images
            \item $\bW$ - Content segmentation map
            \item $L_{max}$ - Number of resolution layers
            \item $n_1,n_2, ~\ldots~,n_m$ - patch sizes
            \item $d_1,d_2,~\ldots~,d_m$ - $\Omega$ subsampling gaps
            \item $I_{IRLS}$ - number of IRLS iterations (chosen as $10$)
            \item $I_{alg}$ - number of update iterations per patch-size
            \item $r$ - robust statistics value to use.
    \end{itemize}

    \item {\bf Initialization:}
    \begin{itemize}
            \item Apply color transfer from $\uS$ to $\uC$
            \item Build the Gaussian pyramids of $\uC$, $\uS$, and $\bW$
            \item Optional (for fast and approximate NN): Prepare the patches from the style image for each resolution layer and each patch size as tree structure
                \item Initialize $\uX = \bD_L^C \uC + V$, where $V \sim {\cal N}(0,50)$
    \end{itemize}

    \item {\bf Loop Over Scales:} For $L=L_{max}, ~\ldots~, 1$ do:
    \item ~~~{\bf Loop Over Patch-Sizes:} For $n=n_1, ~\ldots~, n_m$, minimize $E_{L,n}$ w.r.t.$\uX$, by
    \item ~~~~~~{\bf Iterate:} For $k=1,2 ~\ldots~, I_{alg}$ do:

               \begin{enumerate}
                  \item \emph{Patch Matching}: For the current image $\uX$, match NN from the style image by solving (\ref{eq:NNtask}). The style patches should be of the same size, taken from the corresponding resolution layer.
                  \item \emph{Robust Aggregation}: Compute ${\tilde \uX}$ as the robust patch aggregation result, by solving (\ref{eq:IRLStask}) using $I_{IRLS}$ iterations.
                  \item \emph{Content Fusion}: Combine the content image $\bD_L^C \uC$ to ${\tilde \uX}$ using Equation (\ref{eq:MERGEtask}) to obtain the updated $\uX$.
                  \item \emph{Color Transfer}: Apply color-transfer from $\uS$ to $\uX$.
                  \item \emph{Denoise}: Filter the obtained $\uX$ by the domain-transform.
               \end{enumerate}

    \item ~~~{\bf Scale-Up:} As we move from one resolution layer to the next, scale-up $\uX$.

    \item {\bf Result:} The output of the above algorithm is $\uX$.

    \end{description}
\end{minipage}
\\ \hline
\end{tabular}
\\ \vspace{0.1in}
\caption{The overall Style-Transfer Algorithm.} \label{fig:algo}
\end{center}
\end{figure}

%%%%%%%%%%%%%%%%%%%%%%%%%%%%%%%%%%%%%%%%%%%%%%%%%%%%%%%%%%%%%%%%%%%%%%%%%%%%%%%%
%%%%%%%%%%%%%%%%%%%%%%%%%%%%%%%%%%%%%%%%%%%%%%%%%%%%%%%%%%%%%%%%%%%%%%%%%%%%%%%%

\section{Results}\label{sec:Results}

In this section we summarize several key experiments that illustrate the kind of results expected from the proposed algorithm., and a closer look into the algorithm's ingredients. All the results shown in this section use the following baseline parameter settings, unless said otherwise:
\begin{itemize}
\item All images are of size $400\times 400$ pixels.
\item The patch-sizes are [$33$, $21$, $13$, and $9$].
\item The sub-sampling gaps are [$28$, $18$, $8$, and $5$].
\item The pyramids built have $L_{max}=3$ resolution layers.
\item The robust fusion uses $r=0.8$.
\item Per each patch we apply $I_{alg}=3$ iterations.
\end{itemize}
We recommend to look at the figures of this paper on screen, and zoom-in on the images shown, in order to see fine details and how they are treated.

\subsection{Style-Transfer Examples}

We start by presenting a set of successful examples of style-transfer applied on content images containing faces. Common to all these results is the fact that the segmentation applied was Face-Based. We refrain from showing the segmentation maps in order to save space, and since they are self-evident from the results obtained -- regions segmented as foreground are those in which the final image preserves the content.

\myfigJPGFive{FaceSegment}{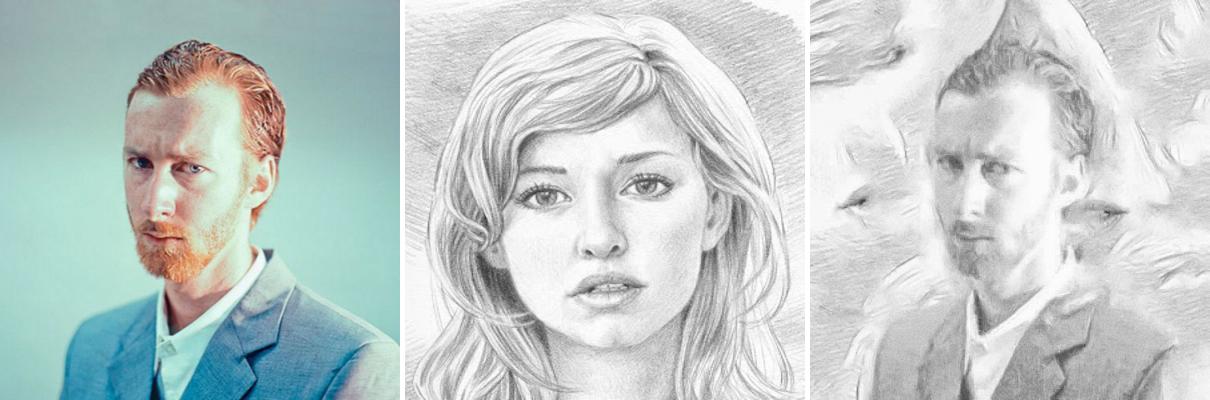}{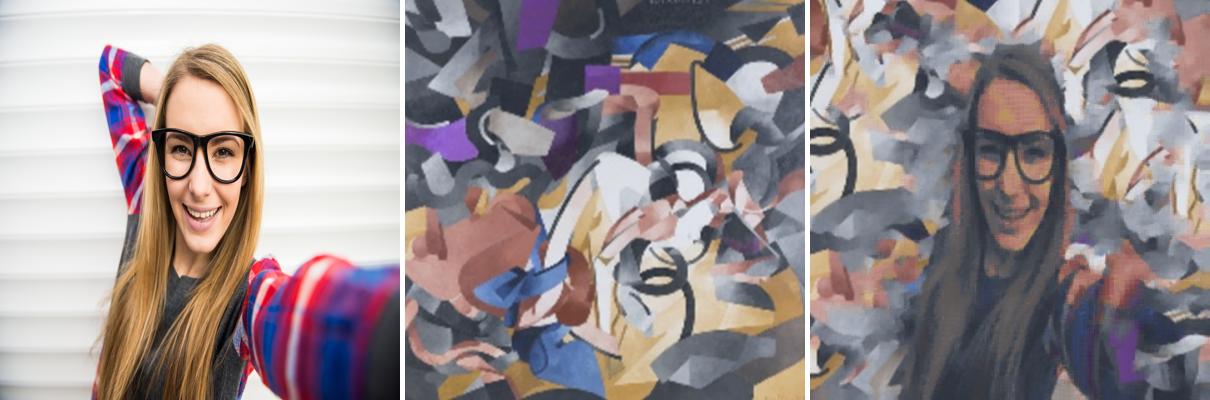}{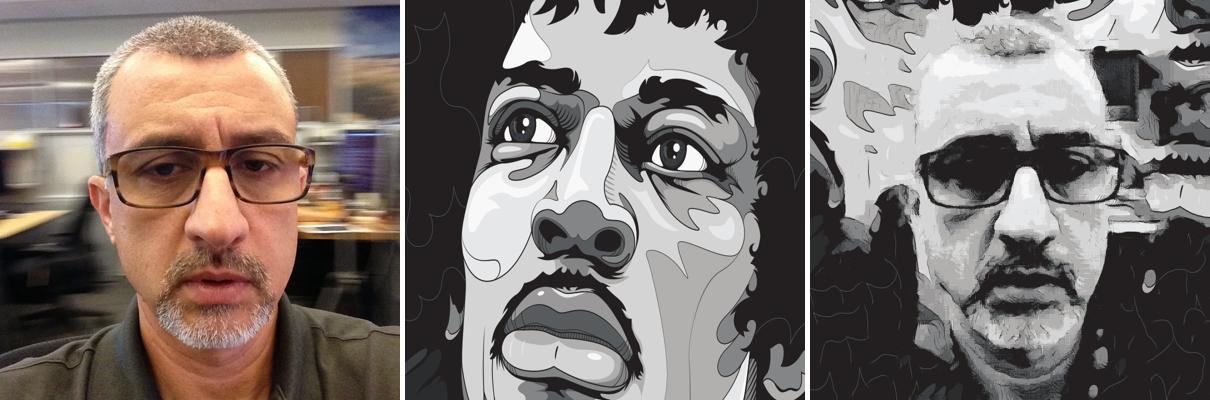}{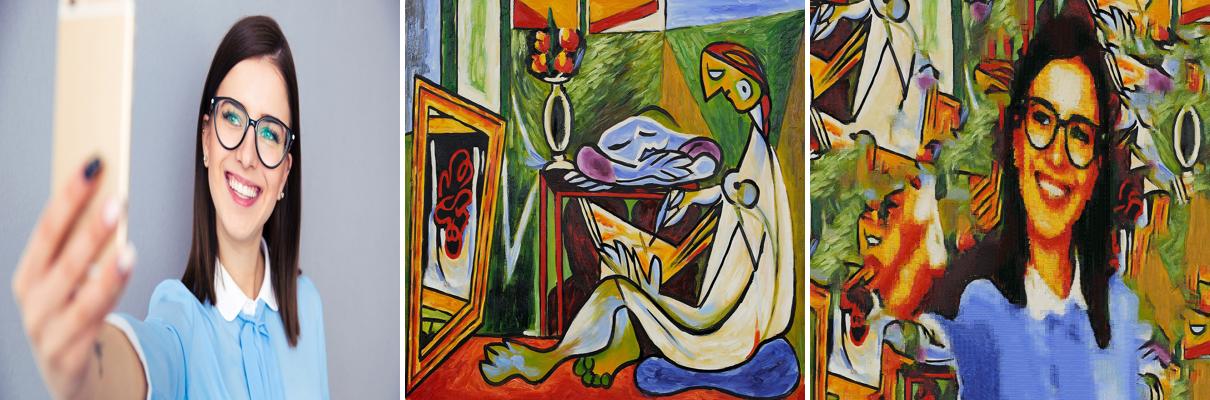}
{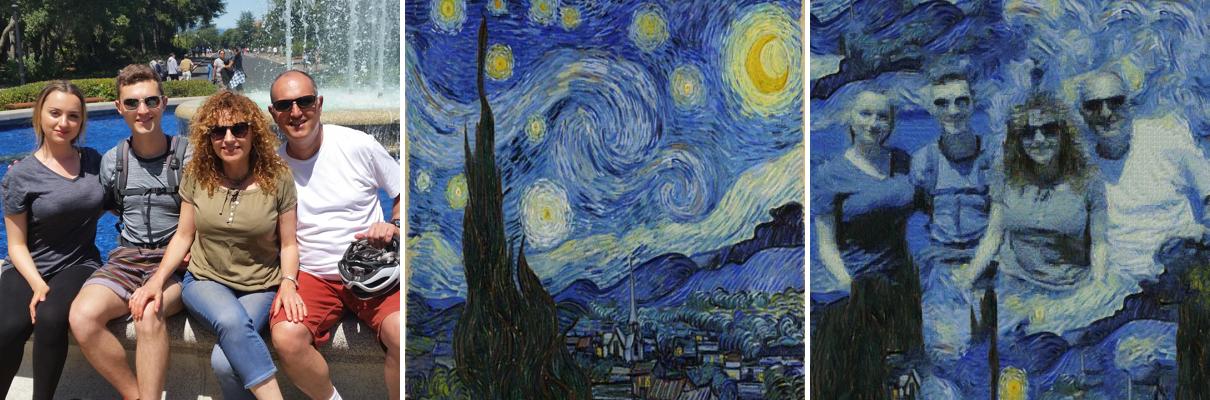}{1.6}{4.8}{{\bf Style-Transfer via Face-Segmentation:} Content image (left), Style image (middle), and the style-transfer result obtained (right). All these images contain faces, and the segmentation applied was Face-Based. }

Few comments are in order with regard to these examples:
\begin{itemize}
\item Notice that the first and third results contain pieces of the face that appear in the style image. Indeed, this effect is more pronounced in the results shown in the Auxiliary part. As commented earlier, these regions in the style images are permitted (along our view) to be used, as they are part of the 'style' definition.
\item In the third example the bottom part is too-dark, and this is a by-product of our palette-transfer applied, which forces the same proportions of dark region in the style and the outcome images.
\item In the fourth example, the hand and the cell-phone vanish, and this is due to the segmentation which decided to mark these regions as background. The same affect takes place in the second example, where the right hand vanishes.
\item Lastly, observe that the segmented content appears almost as is, with slight modifications. If we have chosen to stop the algorithm at a stage in which bigger patches are matched, this would have been different. We will come back to this matter towards the end of this section, when we dive into the intermediate results obtained.
\end{itemize}

We move to show additional style-transfer results, this time on images that were segmented using the Edge-Based method. Apart from this difference, we also added another patch-size of $5\times 5$ to the algorithm's process, used with gap of $3$, in order to get slightly more refined results in content areas.

\myfigJPGFive{EdgeSegment}{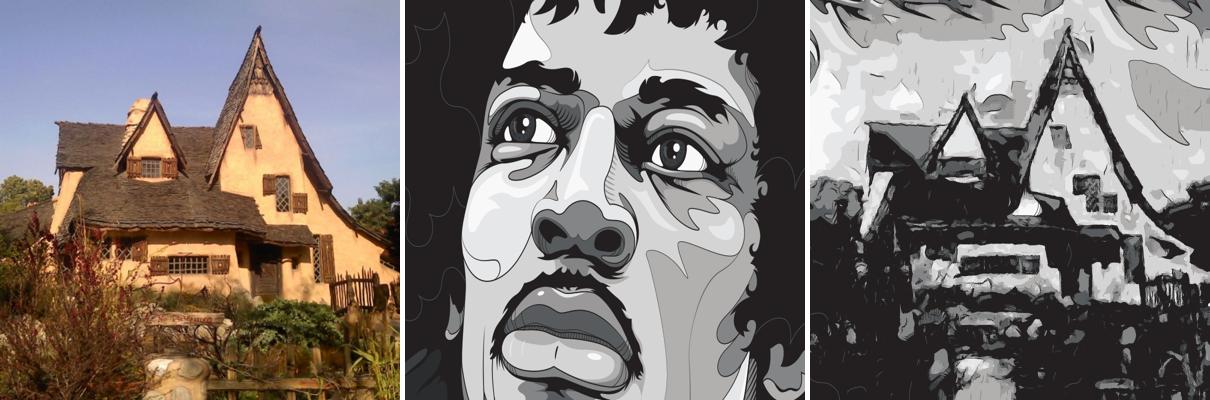}{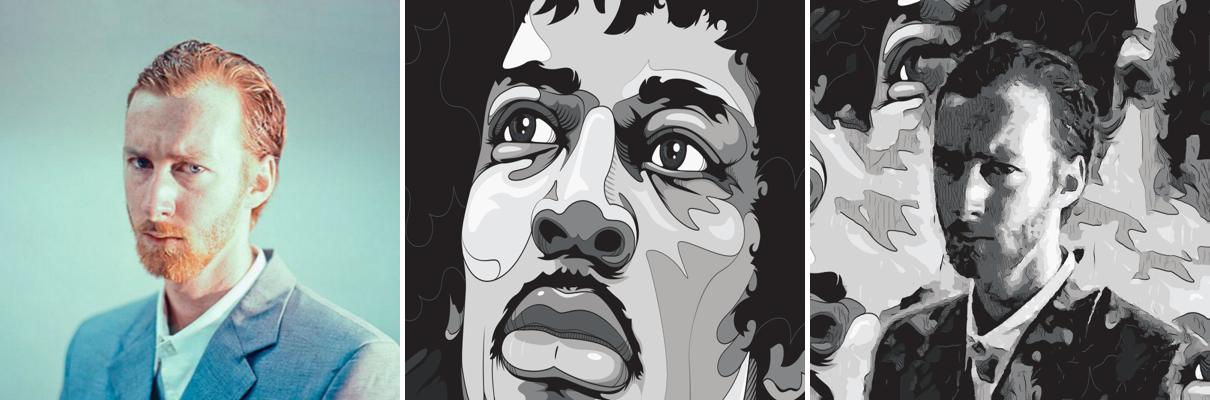}{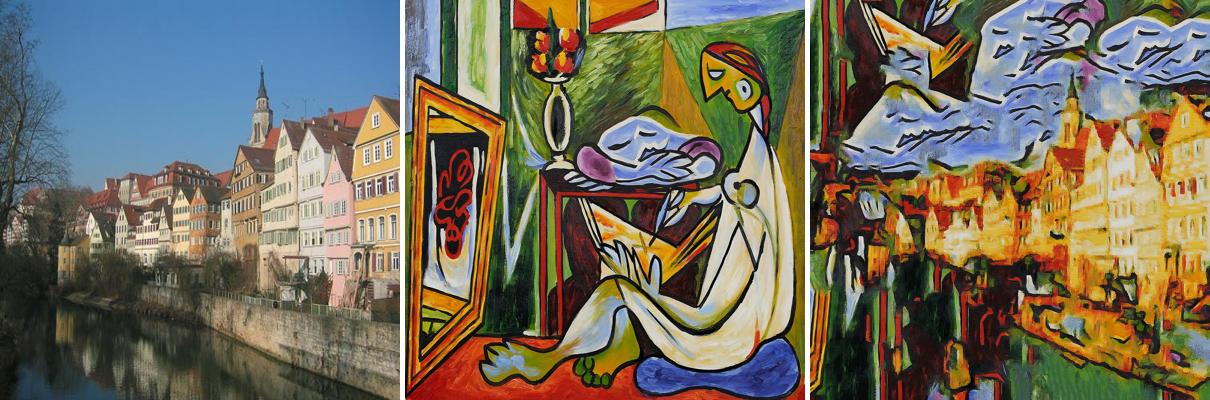}{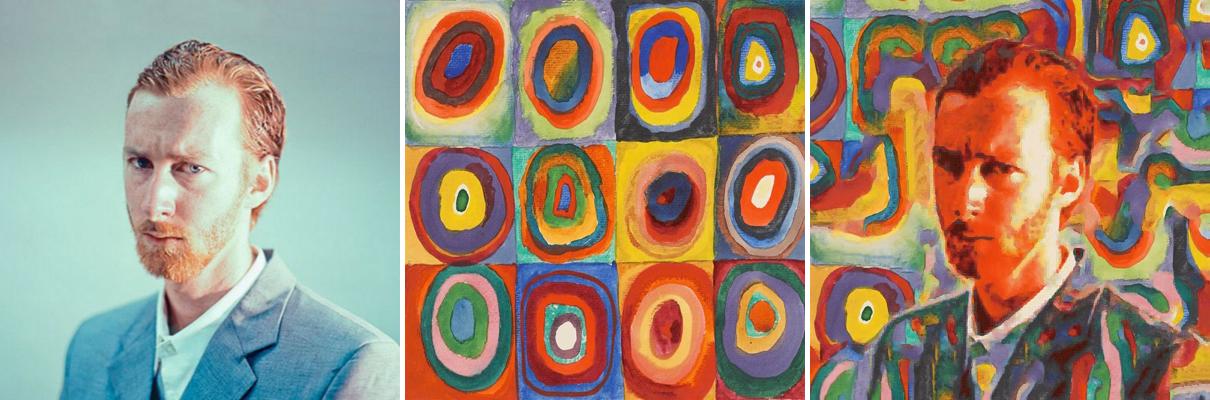}
{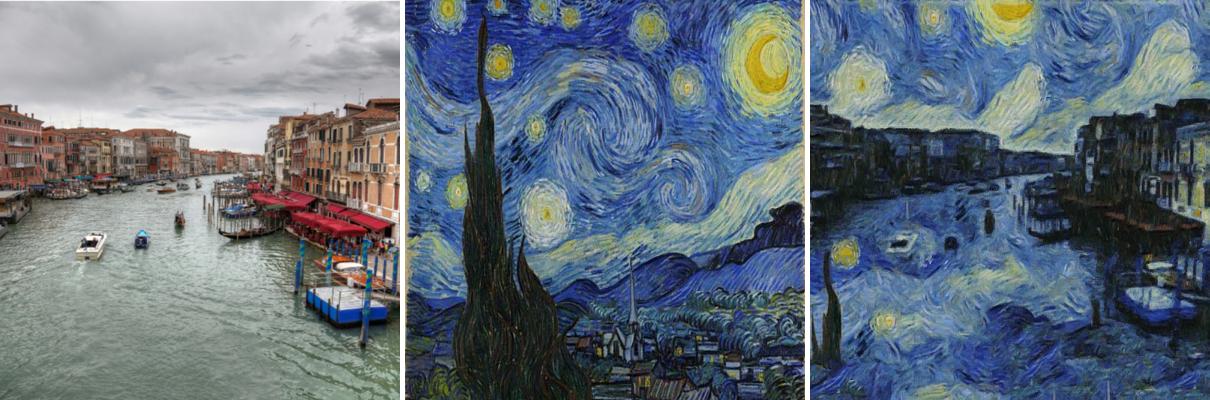}{1.6}{4.8}{{\bf Style-Transfer via Edge-Segmentation:} Content image (left), Style image (middle), and the style-transfer result obtained (right). All these images were segmented using the Edge-Based method.}

Overall, the results obtained in these two experiments provide an acceptable style-transfer effect, and as claimed earlier, the quality obtained is competitive with state-of-the-art. In background regions (based on the segmentation), the result is allowed to hallucinate while being remotely related to the original content, while in the foreground regions the content is, to large extent, preserved.

\subsection{No-Segmentation Results}

We mentioned earlier in Section \ref{sec:Algorithm} that one could apply the proposed algorithm without segmentation, by setting $\bW$ to a constant value. We provide below several such results that correspond a relatively successful transfer. We note that these are the better looking cases, and in other situations the outcome is not visually pleasing (as will be shown later on). Another note is that the success of such transfer depends on the content image (not having too many details), and on the style image (being based on localized elements rather than global brush strokes).

\myfigJPGFour{NoSegment}{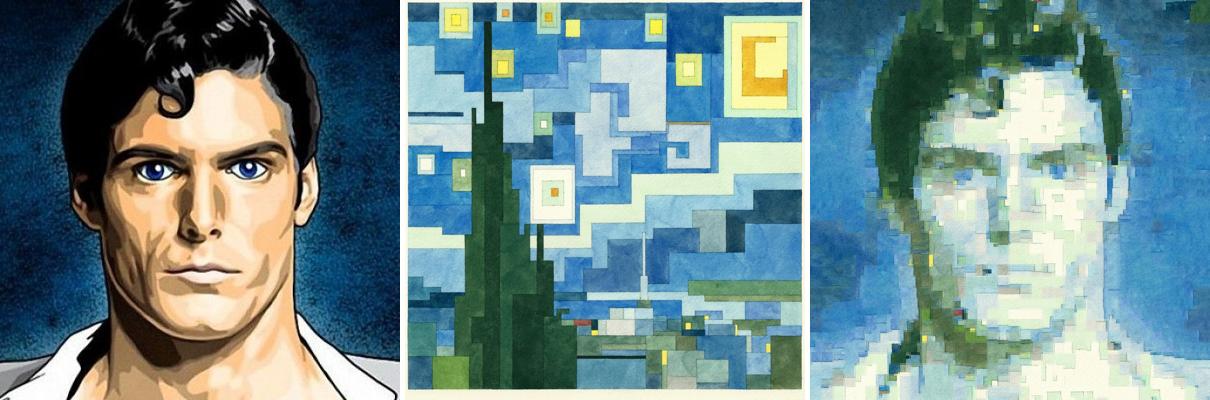}{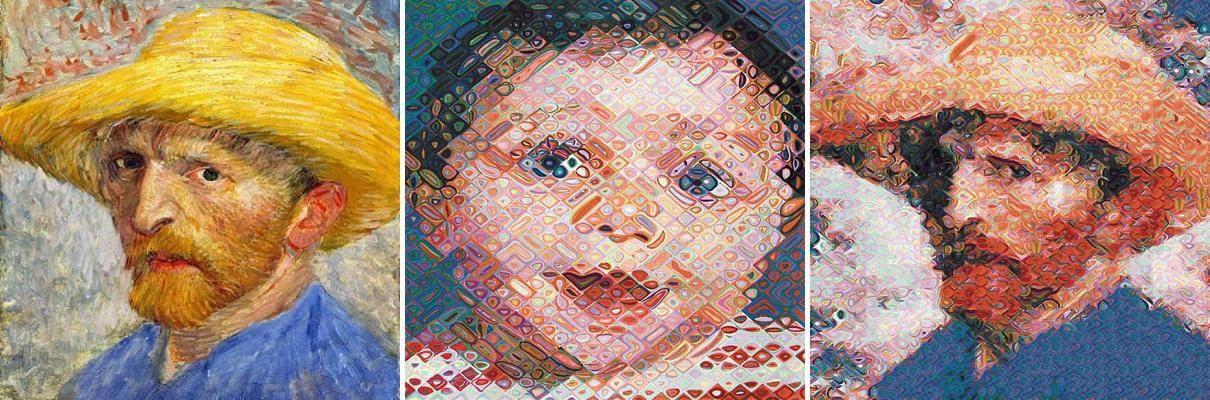}{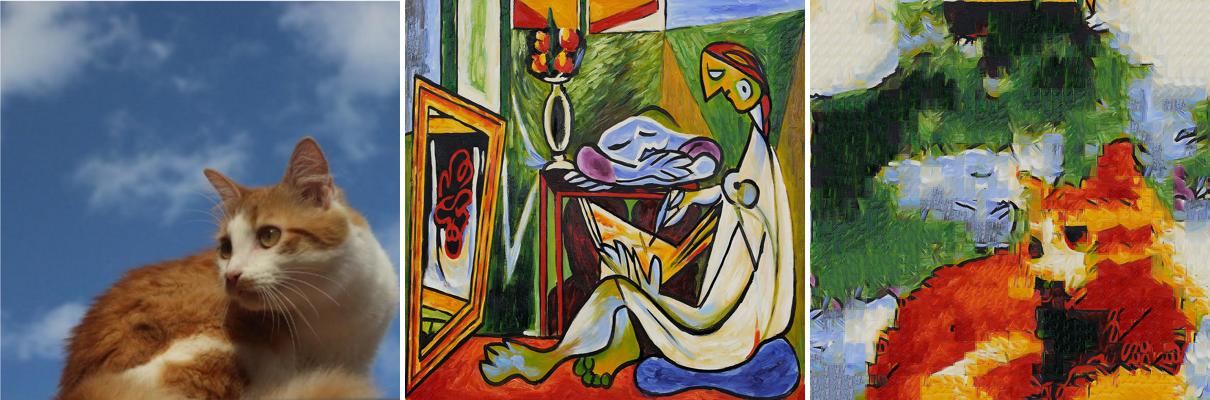}
{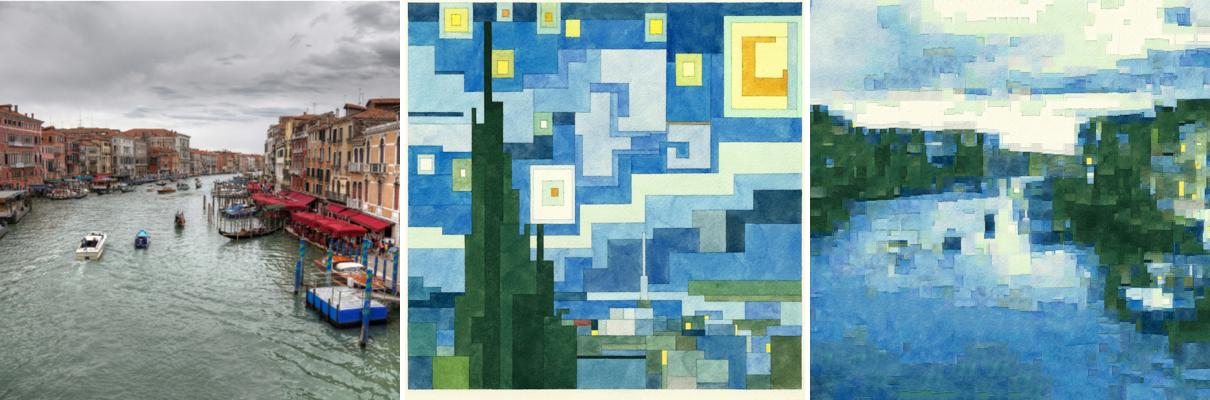}{1.6}{4.8}{{\bf Style-Transfer with No-Segmentation:} Content image (left), Style image (middle), and the style-transfer result obtained (right). All these images were segmented using the Edge-Based method.{\em Zoom-in to better see the behavior of the algorithm on the fine details}.}

\subsection{Tendency to Randomness}

In Figure \ref{fig:Randomness} we show four different results obtained by running the {\em very same algorithm on the very same input pair of content+style and with the very same parameters}. As can be seen, the results are slightly different, owing to the inner randomness that exists within the algorithm. This randomness can be in fact strengthened by allowing to choose more crude approximate nearest neighbor, and by initializing each round by random patch assignments (as proposed in \cite{Kwatra2005}), but we did not explore these directions.

\myfigJPG{Randomness}{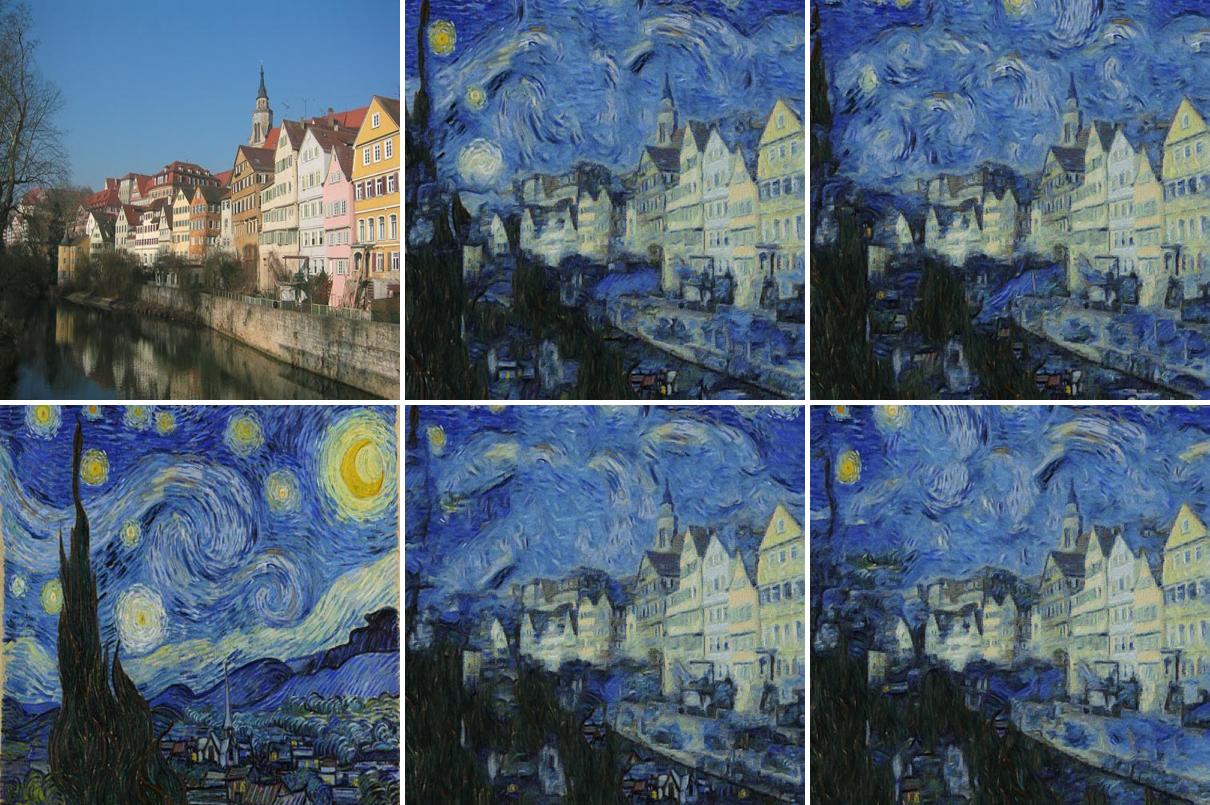}{3.2}{4.8}{{\bf Tendency to Randomness:} Content image (top-left), Style image (bottom-left), and the style-transfer results obtained by running the algorithm four times using the very same settings, and leaning on Edge-Based segmentation. }

\subsection{Comparison to \cite{CNN1} and \cite{Frigo}}

How do these results compare with recent ones that appeared in the literature? We provide a brief comparison to the work by Gatys et. al. \cite{CNN1} and the work by Frigo et. al. \cite{Frigo}. Figure \ref{fig:GatysRes} shows three style-transfer results obtained by Gatys et. al. \cite{CNN1}, and Figure \ref{fig:OursVsGatys} presents the corresponding results obtained by our approach. Similarly, Figure \ref{fig:FrigoRes} shows two style-transfer results obtained by Frigo et. al. \cite{Frigo}, and Figure \ref{fig:OursVsFrigo} presents the corresponding results obtained by our approach.

\myfigJPGThreeHor{GatysRes}{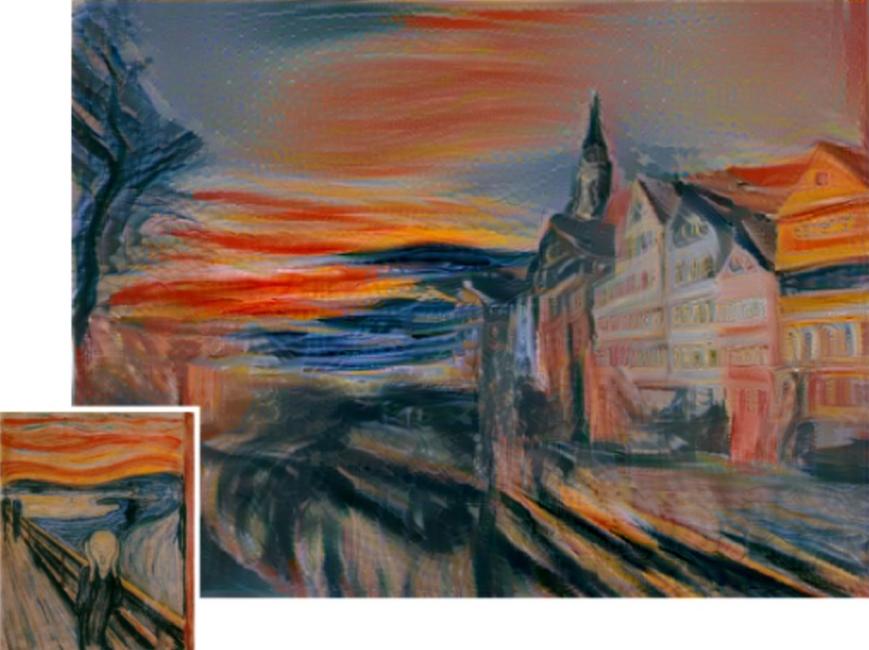}{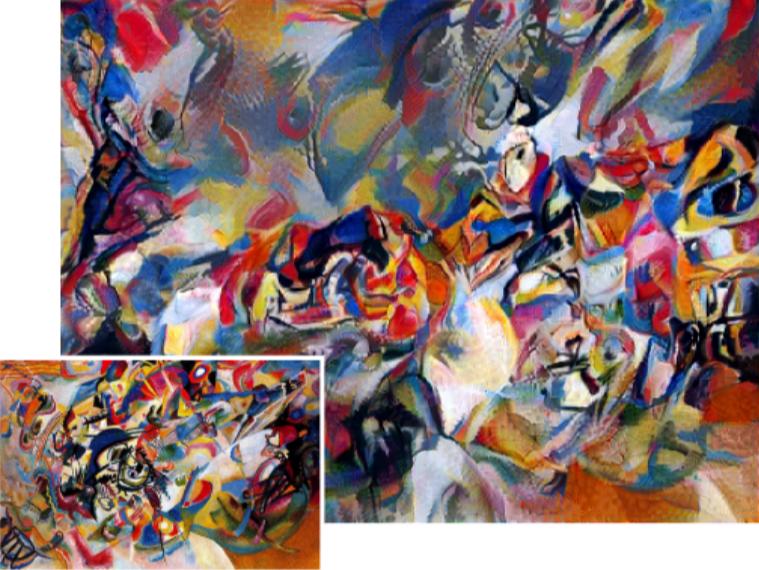}{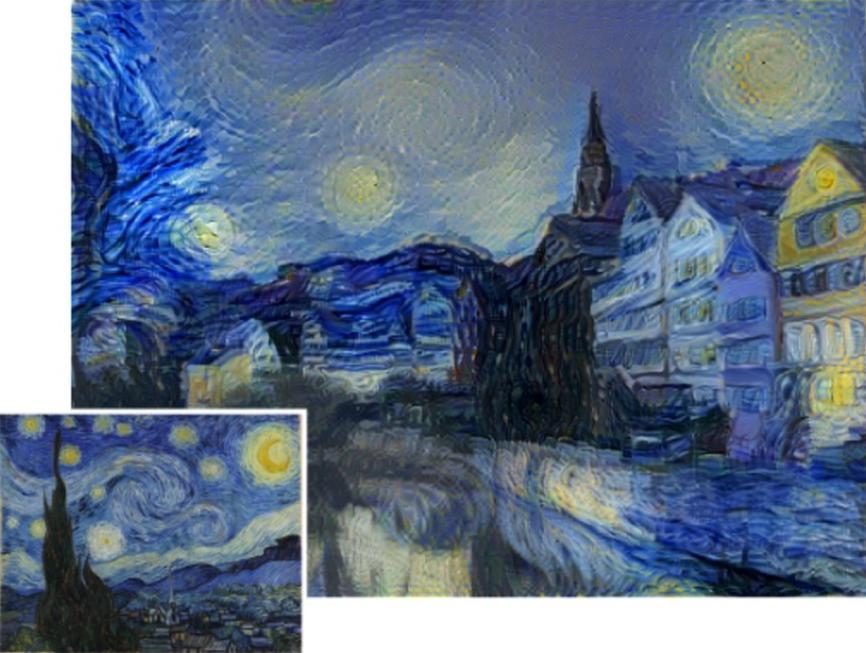}{1.7}{1.7}{{\bf Style-Transfer by Gatys' work \cite{CNN1}:} Three results are shown for three different style images.}

\myfigJPG{OursVsGatys}{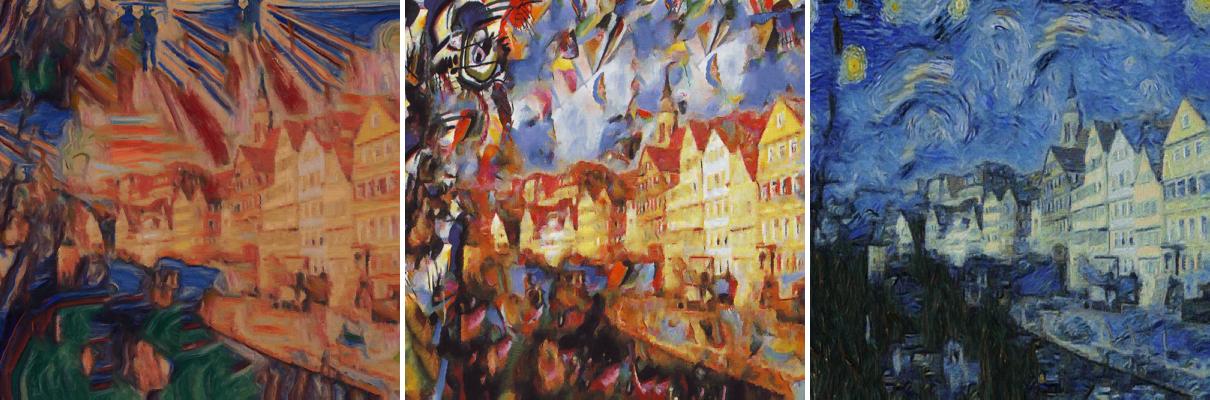}{1.6}{4.8}{{\bf Comparison to Gatys' Results:} These three results are obtained by our algorithm, and should be compared to the images shown in Figure \ref{fig:GatysRes}.}

\myfigJPG{FrigoRes}{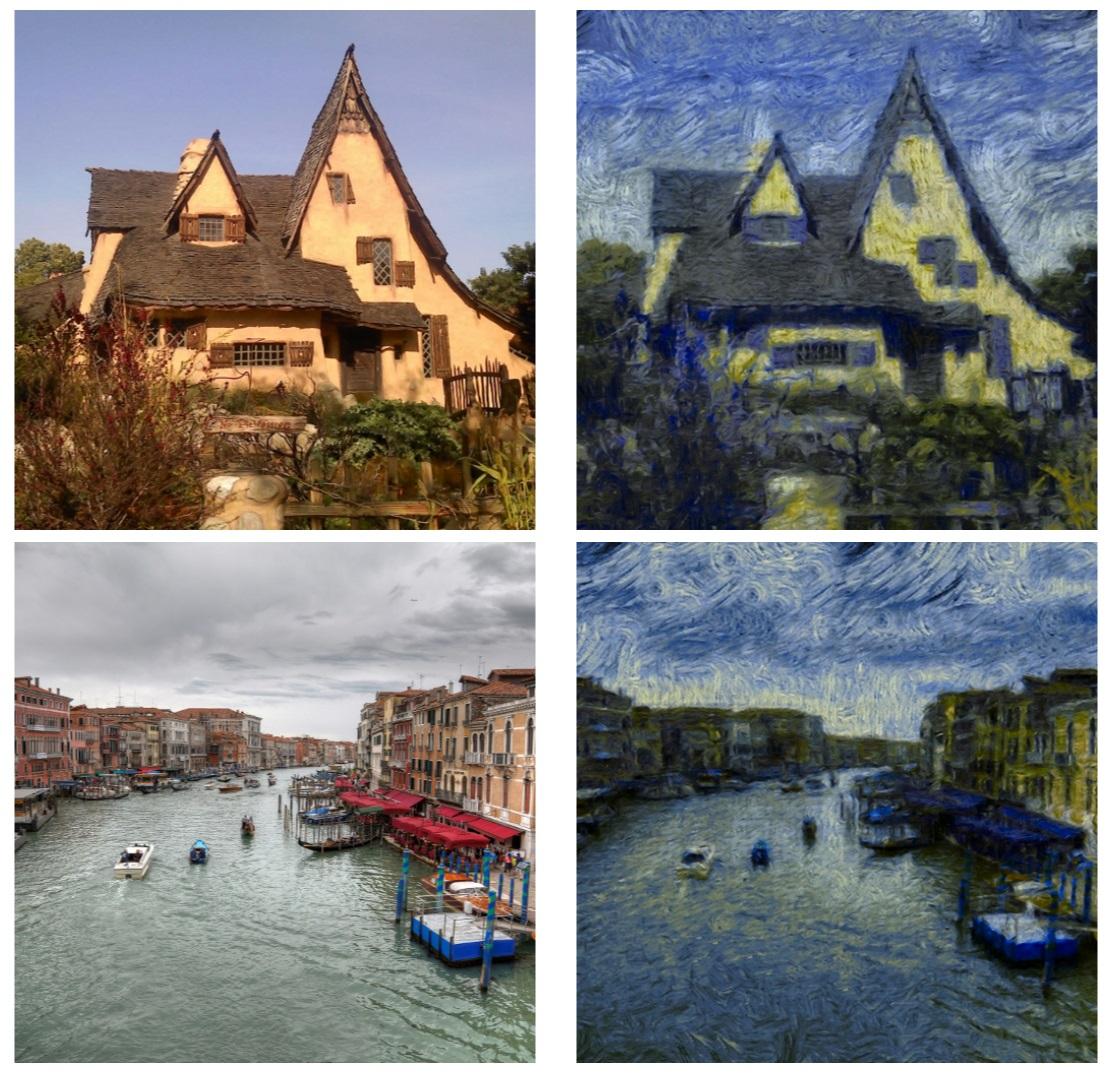}{3.2}{3.2}{{\bf Style-Transfer by Frigo' work \cite{Frigo}:} Two results are shown for two content images transferred by a common style image (Starry-Night).}

\myfigJPG{OursVsFrigo}{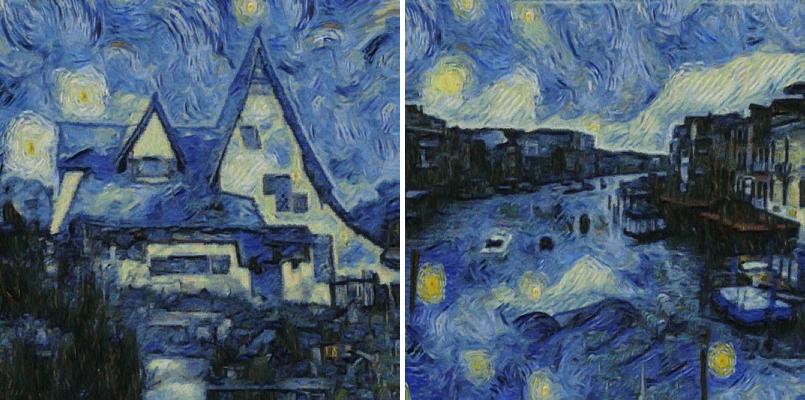}{1.6}{3.2}{{\bf Comparison to Frigo' Results:} These two results are obtained by our algorithm, and should be compared to the images shown in Figure \ref{fig:FrigoRes}.}

Overall, the style-transfer obtained by our method seems to be competitive with Gatys' results, and perhaps slightly inferior to it. In terms of computational complexity, the proposed algorithm is far faster, and does not require special pre-training. In the images that are compared to Gatys' work, the palette transfer plays a key role in the results obtained. The skies in the content image is forced to be bluish, and this poses a limit to the pieces from the style images that are eventually adopted for this region. Another key feature that characterize Gatys' results and is missing from ours are the very long brush-strokes from the style. This is extremely evident in the first and third examples, where such strokes exist. Lastly, we see that our algorithm is more 'conservative' in handling the foreground (segmented) content. All these highlight important directions of future research, with the hope to further strengthen the results obtained by the proposed algorithm.

In comparison to Frigo's work, our images tend to be more vivid, adopting larger and more diverse style ingredients. This is mostly due to the freedom some regions in the image are given in hallucinating their final content.

\subsection{Failure Cases}

Despite the fact that the proposed algorithm is stable and robust, we do get unsatisfactory results on some of the input pairs of content+style. We present several such cases and discuss the reasons for the failures:
\begin{itemize}
\item {\bf Poor Palette-Transfer:} Figure \ref{fig:FailureColor} presents two failure cases in which the initial palette transfer ruined the outcome of the algorithm. In both these cases, the level of brightness and contrast on the objects of interest were set badly, washing-out the important details. The reason for this mistake is obvious - the brightest part in the content image is matched to the brightest colors of the style, and when over-done, the effect is a severe loss of quality.
\item {\bf Poor Segmentation:} Figure \ref{fig:FailureSeg} presents two style-transfers in which the poor segmentation is the reason to the failure. In both cases the segmentation used was the Edge-Based, and it fails to recover the complete faces. As a consequence, regions of the faces were overridden too aggressively by style.
\item {\bf Mismatch Between Content and Style:} Figure \ref{fig:FailureMisMatch} presents two transfer results that failed due to a miss-match between the content and the style images. The heads in the content image are very small, while the core elements of the style are much larger. In such a case, while the outcome may be very rich and interesting, faces are unrecognizable, and most users would consider this as failure.
\item {\bf Poorly Chosen Style Image:} Figure \ref{fig:FailureStyle} presents two failure cases, caused by the poor choice of the style image. A fundamental question that has been overlooked in the style-transfer literature is when a given style image could be considered as appropriate? One trivial (and partial) answer is that the style image should have a rich palette. Large bright (or dark) regions are likely to ruin the results of the transfer, as in the second example shown. Low number of expressive colors may also have a devastating effect, as happens in the first example.
\item {\bf Poor Results without Segmentation:} Lastly, Figure \cite{fig:FailureNoSeg} shows two results in which no segmentation was applied, and which led to unpleasant outcome. This failure is reflected mostly by the loss of critical details that make these photos recognizable. This is especially important for images containing faces, as in the second example.
\end{itemize}

\myfigJPGTwo{FailureColor}{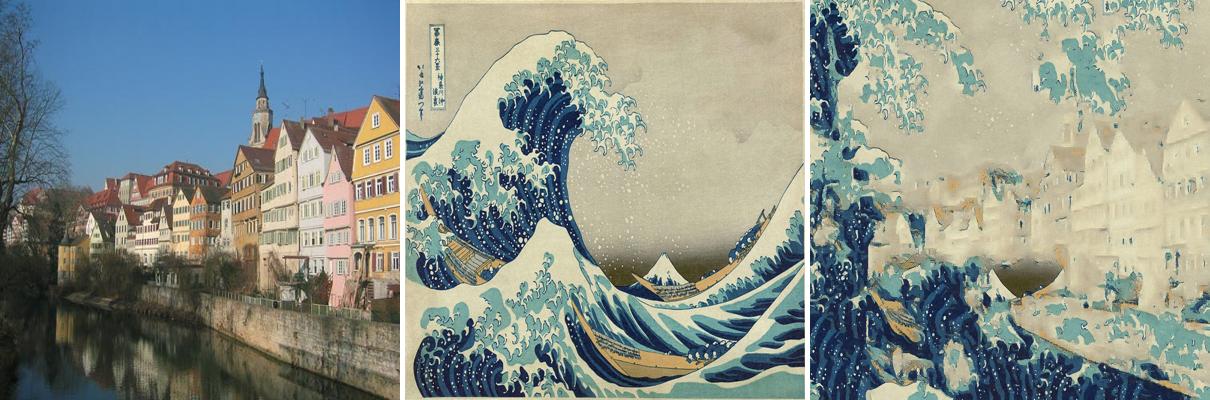}{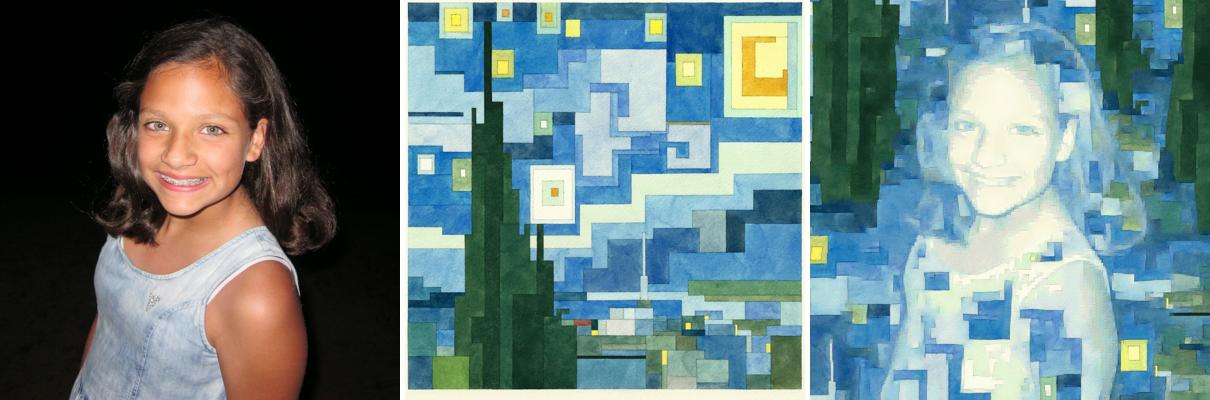}{1.6}{4.8}{{\bf Failure due to Poor Palette Transfer:} These two results are obtained by our algorithm, and their poor quality is mostly due to the poor palette transfer applied within our method.}

\myfigJPGTwo{FailureSeg}{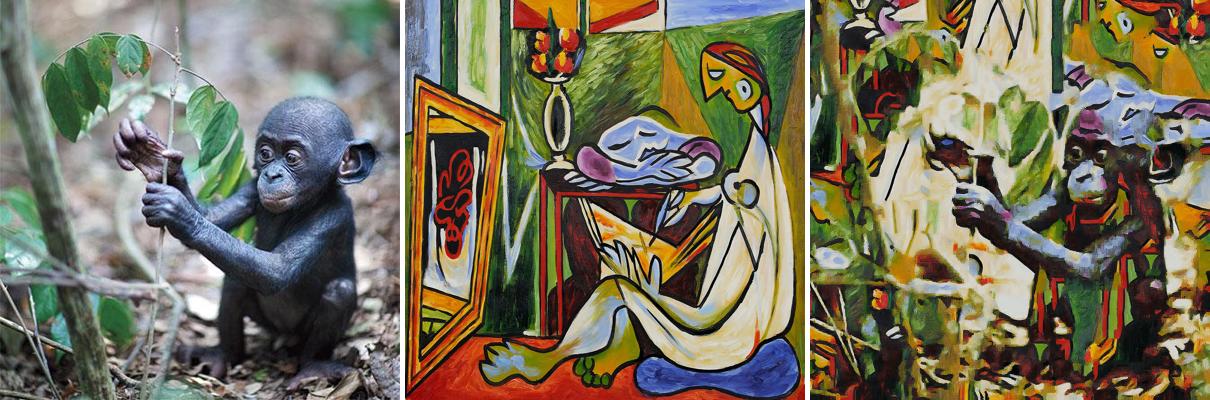}{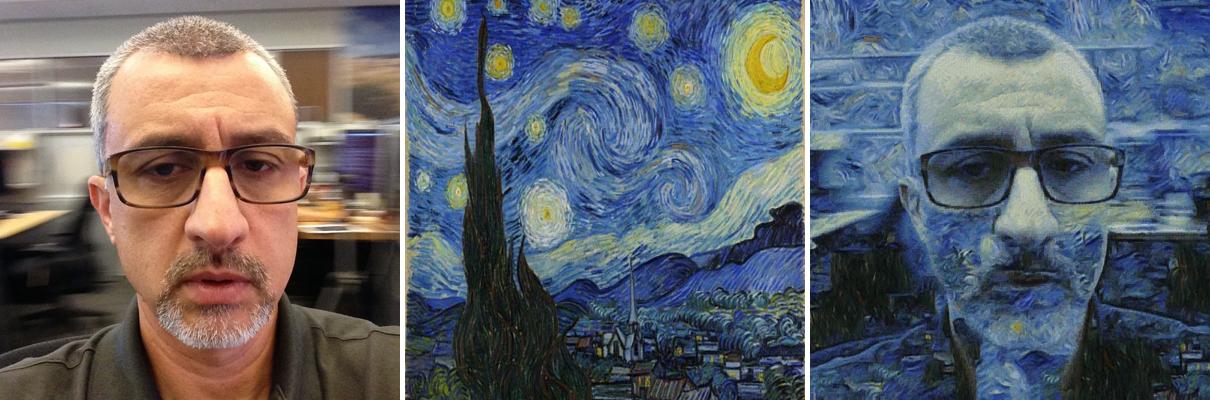}{1.6}{4.8}{{\bf Failure due to Poor Segmentation:} These two results are obtained by our algorithm, and their poor quality is due to the failure of the segmentation algorithm (Edge-Based) to identify the critical face regions.}

\myfigJPGTwo{FailureMisMatch}{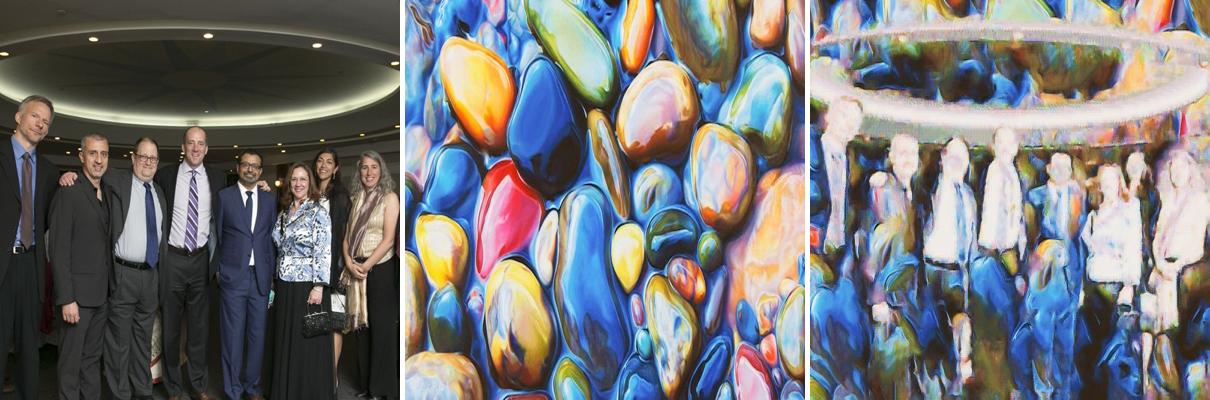}{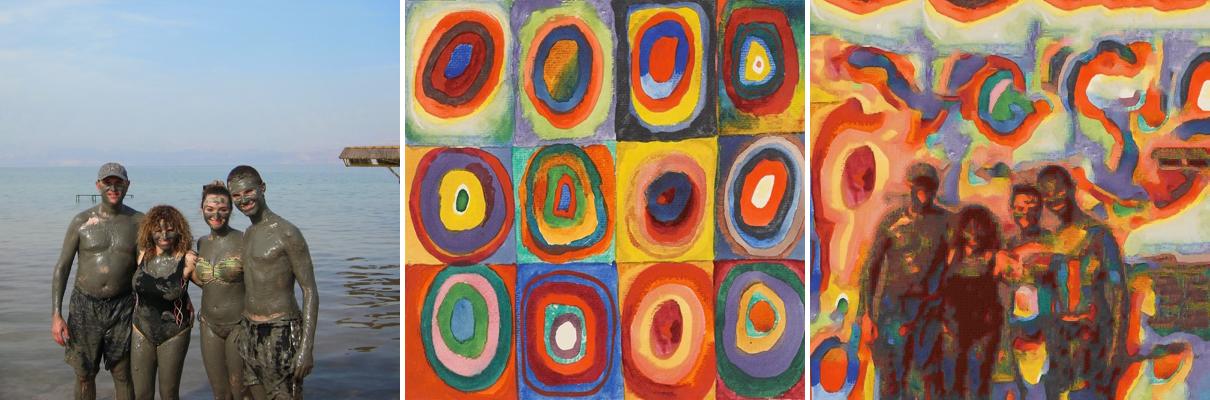}{1.6}{4.8}{{\bf Failure due to Poor Match:} These two results are obtained by our algorithm. There is a fundamental mismatch between the object sizes in the content and the style, which causes loss of critical information.}

\myfigJPGTwo{FailureStyle}{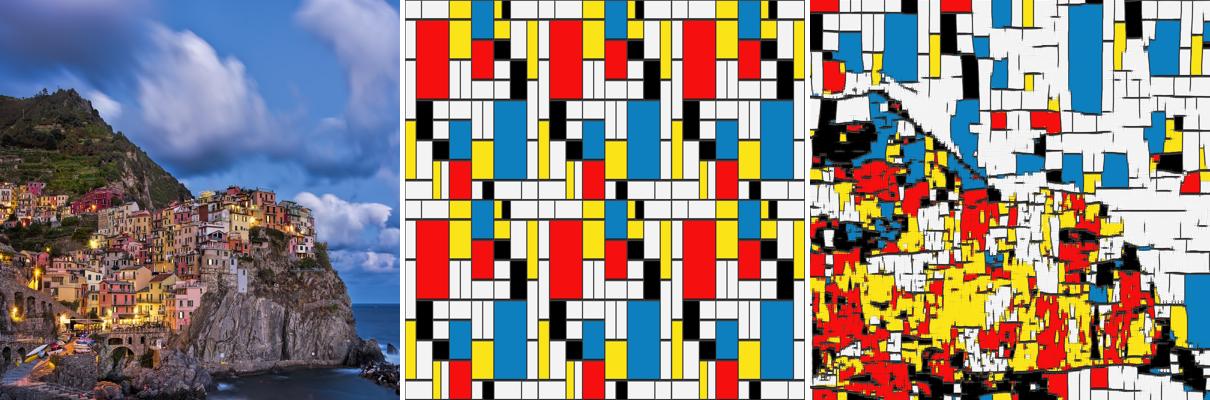}{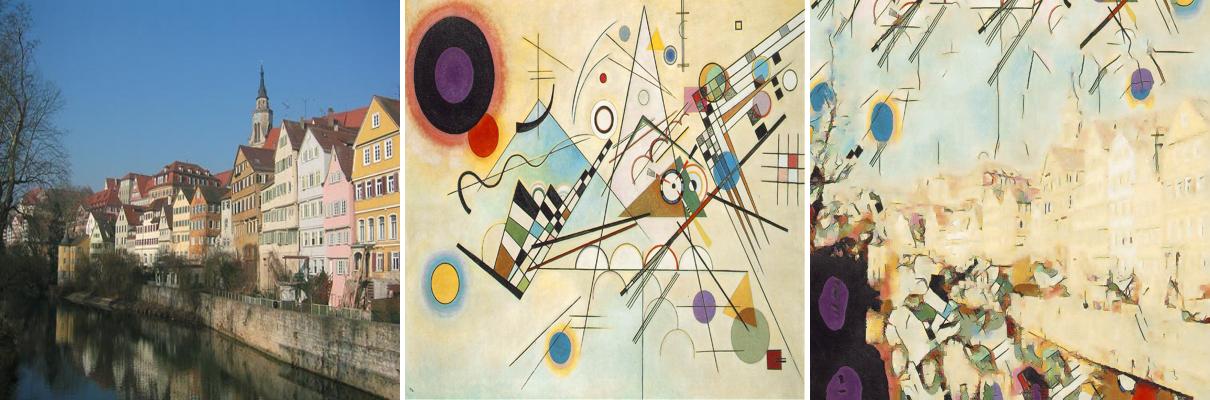}{1.6}{4.8}{{\bf Failure due to Poorly Chosen Style Image:} These two results are obtained by our algorithm, and in both the chosen style image is not rich enough to provide with a proper transfer.}

\myfigJPGTwo{FailureNoSeg}{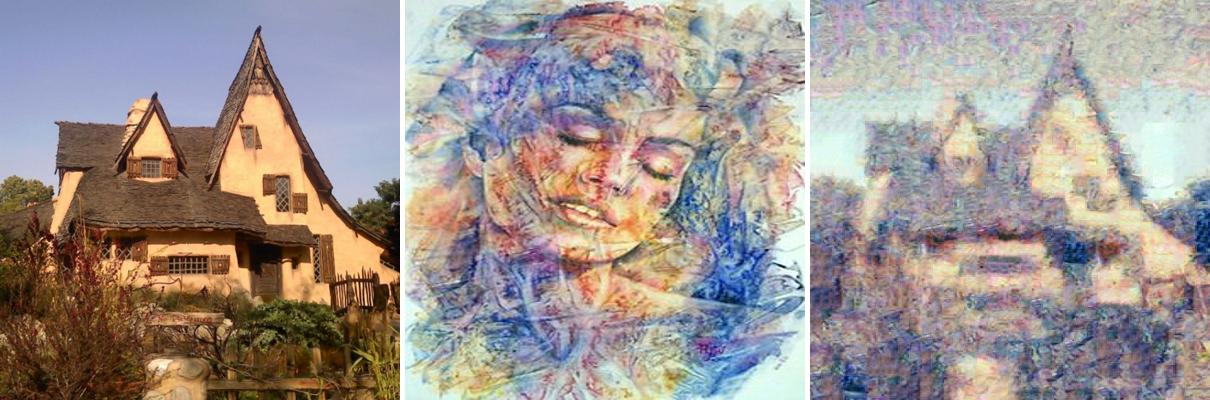}{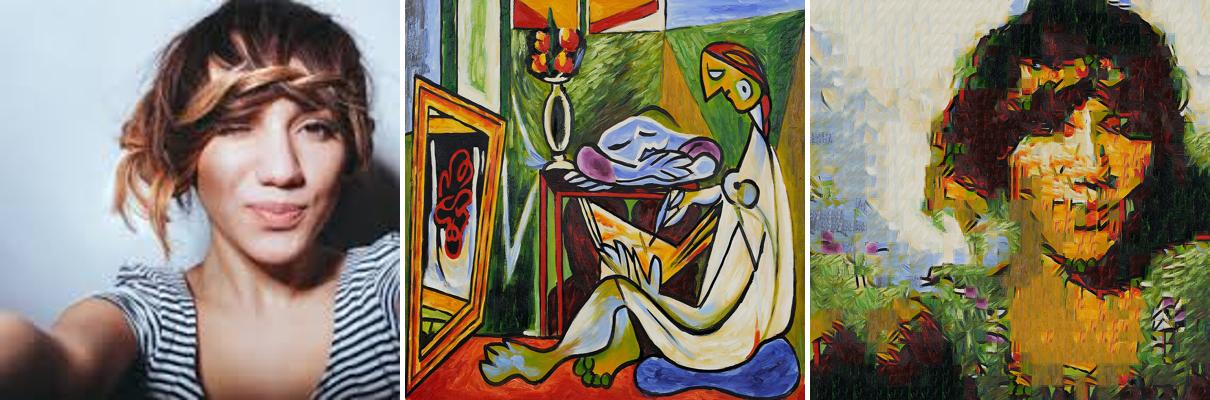}{1.6}{4.8}{{\bf Failure due to No-Segmentation:} These two results are obtained by our algorithm, and in both we applied no segmentation. As can be seen, the important content parts are treated badly and lost.}

\subsection{A Closer view of the Optimization Process}

We turn to briefly present the inner operations of their proposed algorithm, and how the result is incrementally built. Figure \ref{fig:Input} shows the input content- and style-images we start with. The very first step taken is palette-transfer of the content image to match that of the style, and this is shown as well in this figure. Also shown is the content segmentation map to be used within the algorithm.

{\begin{figure}[htbp]
   \begin{center}
     \includegraphics[height=1.6in,width=4.8in]{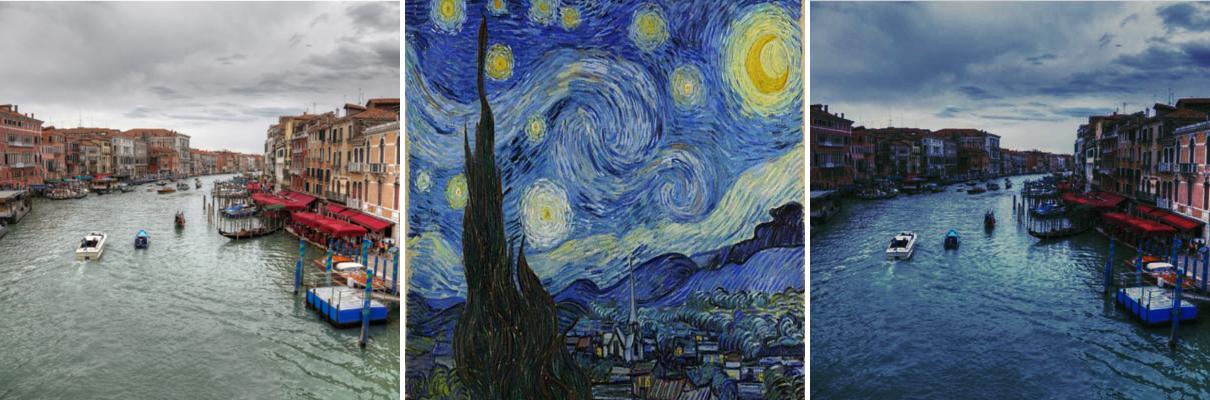}\\
     \includegraphics[height=1.6in,width=1.6in]{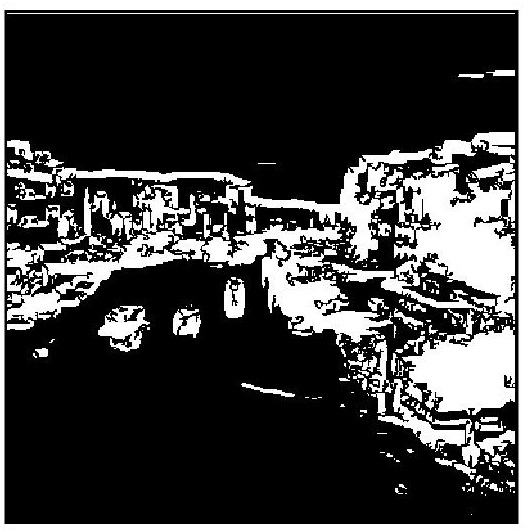}
     \caption{{\bf A Closer view of the process:} Top: The input content- and style-images and the color-transferred content to initiate the algorithm with. Bottom: The edge-based segmentation used by the algorithm.}
     \label{fig:Input}
   \end{center}
\end{figure}}

Figure \ref{fig:Graph} presents the penalty term in Equation (\ref{eq:IRLStask}), which is the core of our optimization process\footnote{We could have shown the full term with the content-match as in (\ref{eq:IRLStaskALL}), but this will obscure some important effects that take place in the different stages of the algorithm.}. First, observe that the entire graph is separated into three parts, each representing a different resolution layer in the pyramidal treatment. The left-most corresponds to images of size $100\times 100$ pixels, the second operates on $200\times 200$ pixels images, and the last (right-most) considers the native image resolution of $400\times 400$ pixels. Each of these is initialized with an image augmented by strong additive Gaussian noise.

Within each layer, we sweep through several patch sizes, from the biggest to the smallest. In each such case, we target an optimization objective function $E_{L,n}$, where $L$ is the resolution layer and $n$ is the patch-size. For each of these, we apply several steps:
\begin{enumerate}
\item {Patch-matching:} Given the temporary image $\uX$  we seek the nearest neighbors for each patch on a sampled grid $\Omega$. Since we are using approximated NN, we are not necessarily getting a descent in the function height in this stage.
\item {Patch-aggregation:} Given the found NN patches, we update the image $\uX$ using IRLS, and this causes a sharp drop in the function value. One can see that the choice to run $110$ iterations seems to be too generous, and $2-3$ would have been sufficient. This leads to the image ${\tilde \uX}$.
\item {Content-enforcement:} Given the image ${\tilde \uX}$, we merge the content to it in chosen areas based on the segmentation map, and this necessarily leads to an increase in the function height. This gives the image ${\hat \uX}$.
\item {Palette-transfer:} The final step is to enforce the style palette on ${\hat \uX}$, which is typically accompanied by a further increase in the function height.
\end{enumerate}

\noindent Observe that in the very last stage (when handling $E_{1,7}$, we omit the content-enforcement and the palette-transfer stages, since we aim to get delicate processing of the content regions as well.

{\begin{figure}[htbp]
   \begin{center}
     \includegraphics[height=2in,width=7in]{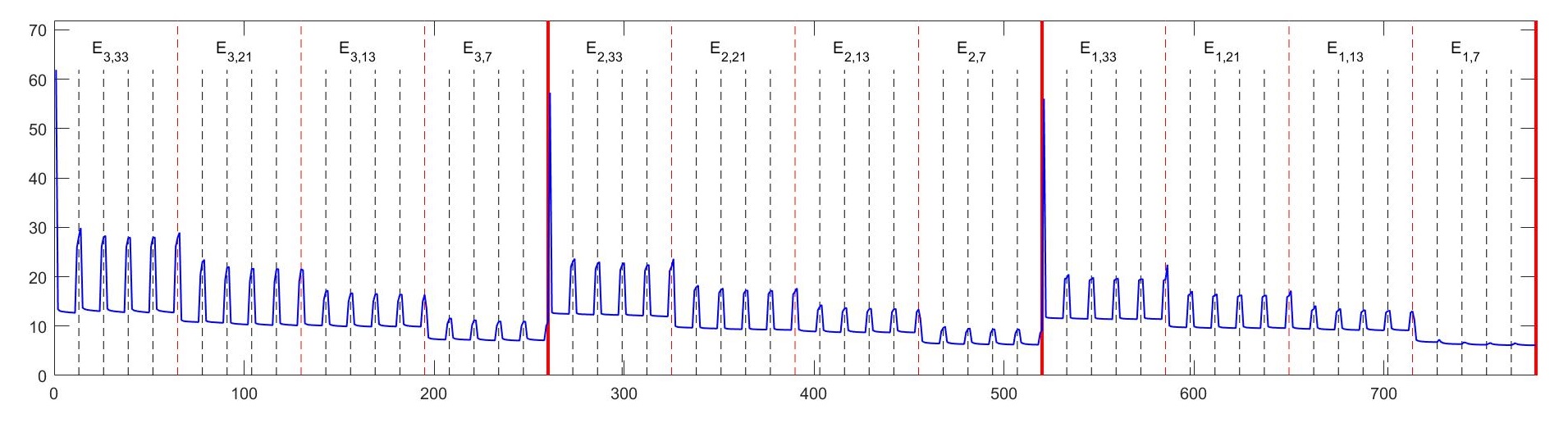}
     \caption{{\bf A Closer view of the process:} The penalty function in Equation (\ref{eq:IRLStask}) as a function of the iterations. Notice the separation of the whole graph into the three resolution levels, and in each the separation to different patch-sizes. Each part is marked with the penalty function it is targeting.}
     \label{fig:Graph}
   \end{center}
\end{figure}}

How do the images look as the algorithm evolves? Figure \ref{fig:Process_Images} aims to answer this by presenting for each resolution layer the initialization image, the images after each patch-aggregation step, and the images obtained after the content-enforcement and the palette-transfer. As can be observed, the image is refined and improved as we go up the pyramid and down the patch-size, and one could envision omitting few of these steps without much harm to the final image.

{\begin{figure}[htbp]
   \begin{center}
     \includegraphics[height=0.3in,width=0.3in]{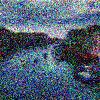}\\ \vspace{0.05in}    \includegraphics[height=0.3in,width=0.3in]{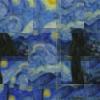}
     \includegraphics[height=0.3in,width=0.3in]{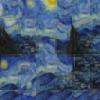}
     \includegraphics[height=0.3in,width=0.3in]{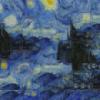}
     \includegraphics[height=0.3in,width=0.3in]{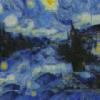}\\ \vspace{0.05in}
     \includegraphics[height=0.3in,width=0.3in]{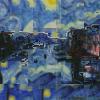}
     \includegraphics[height=0.3in,width=0.3in]{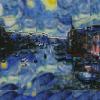}
     \includegraphics[height=0.3in,width=0.3in]{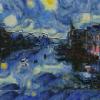}
     \includegraphics[height=0.3in,width=0.3in]{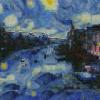}\\ \vspace{0.05in}
     \includegraphics[height=0.6in,width=0.6in]{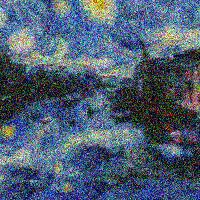}\\ \vspace{0.05in}
     \includegraphics[height=0.6in,width=0.6in]{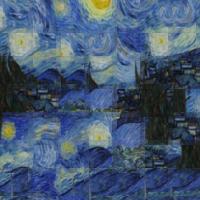}
     \includegraphics[height=0.6in,width=0.6in]{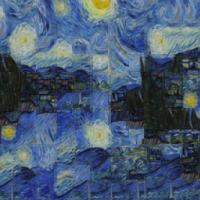}
     \includegraphics[height=0.6in,width=0.6in]{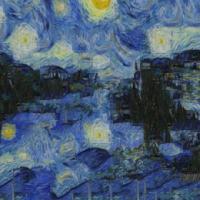}
     \includegraphics[height=0.6in,width=0.6in]{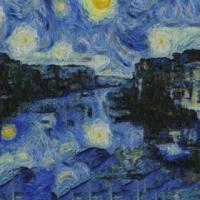}\\ \vspace{0.05in}
     \includegraphics[height=0.6in,width=0.6in]{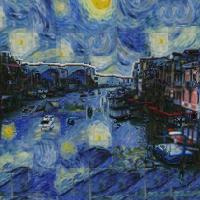}
     \includegraphics[height=0.6in,width=0.6in]{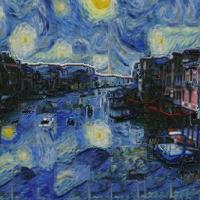}
     \includegraphics[height=0.6in,width=0.6in]{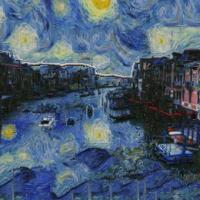}
     \includegraphics[height=0.6in,width=0.6in]{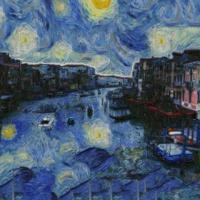}\\ \vspace{0.05in}
     \includegraphics[height=1.2in,width=1.2in]{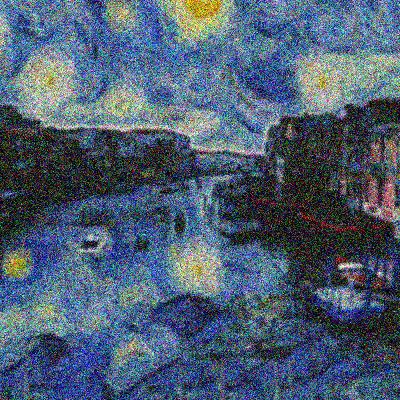}\\ \vspace{0.05in}
     \includegraphics[height=1.2in,width=1.2in]{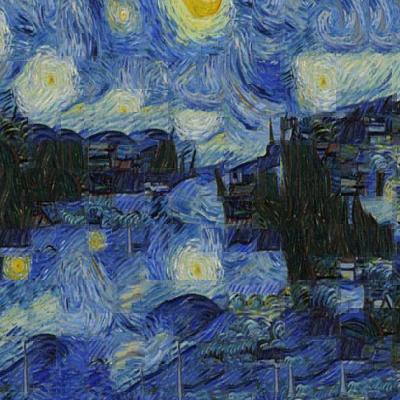}
     \includegraphics[height=1.2in,width=1.2in]{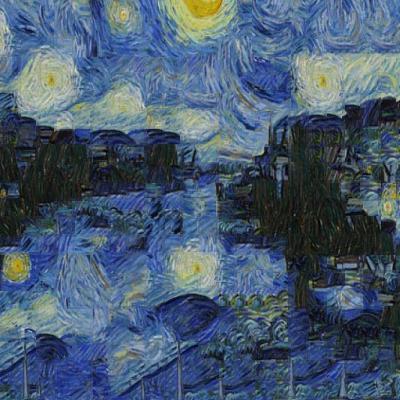}
     \includegraphics[height=1.2in,width=1.2in]{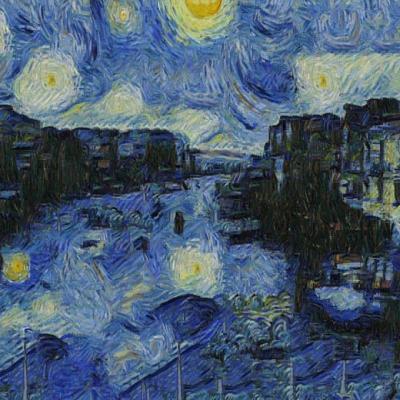}
     \includegraphics[height=1.2in,width=1.2in]{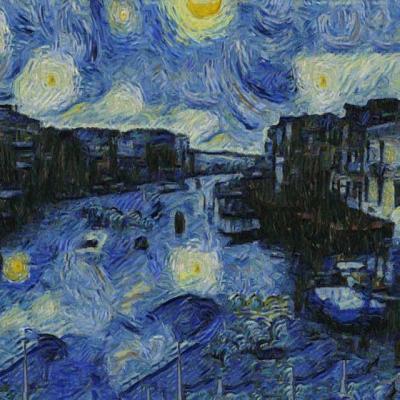}\\ \vspace{0.05in}
     \includegraphics[height=1.2in,width=1.2in]{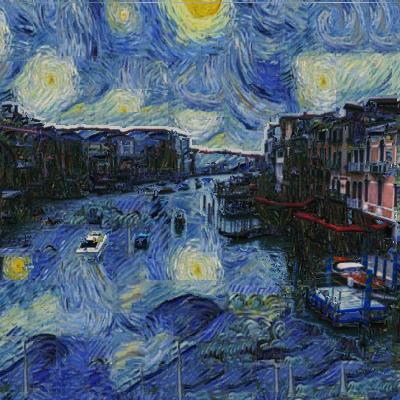}
     \includegraphics[height=1.2in,width=1.2in]{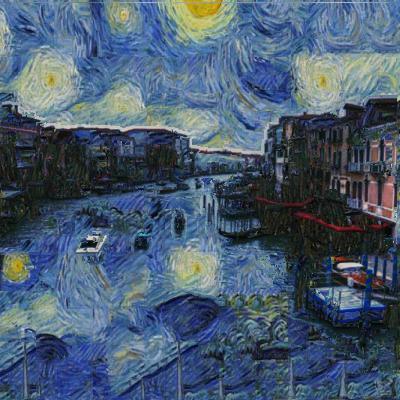}
     \includegraphics[height=1.2in,width=1.2in]{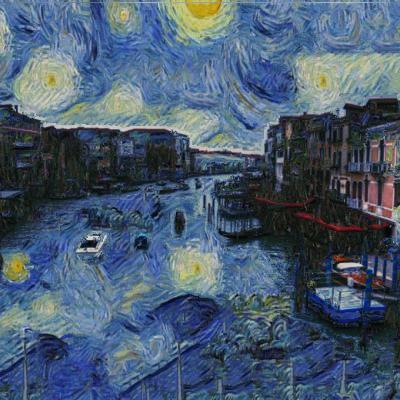}
     \includegraphics[height=1.2in,width=1.2in]{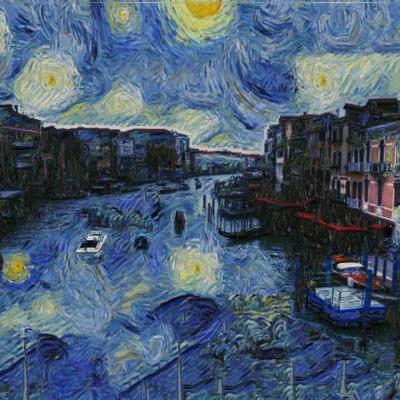}
     \caption{{\bf A Closer view of the process:} The intermediate results obtained in each of the three resolution layers ($100\times 100$, $200\times 200$, and $400\times 400$ pixels) results. In each group we present the initialization image as augmented with noise (first row), the set of images ${\tilde \uX}$ (second row), and their fusion with the segmented content, ${\hat \uX}$ (third row). The output of this process is the last ${\tilde \uX}$ image (without merging the content).}
     \label{fig:Process_Images}
   \end{center}
\end{figure}}

\subsection{Parameters and their Influence}

We conclude the presentation of experimental results with demonstration of the influence of the key parameters that control our algorithm -- the pyramid height $L$, the patch-sizes to use $n$, and their overlap, $d$.

Figure \ref{fig:Leffect} shows the same content+style images, treated by our algorithm (using Edge-Based segmentation) while varying the depth of the pyramid, $L$. As can be clearly seen, the higher the pyramid, the larger are the portions adopted from the style. When $L$ is too high ($4$ in our case), we get a richer outcome, but at the risk of copying complete pieces from the style image, instead of hallucinating new combinations. On the other extreme of $L=1$, the result is more conservative, leaning on the common and local brush-strokes, being somewhat similar to the results by \cite{Frigo}.

\myfigJPGTwo{Leffect}{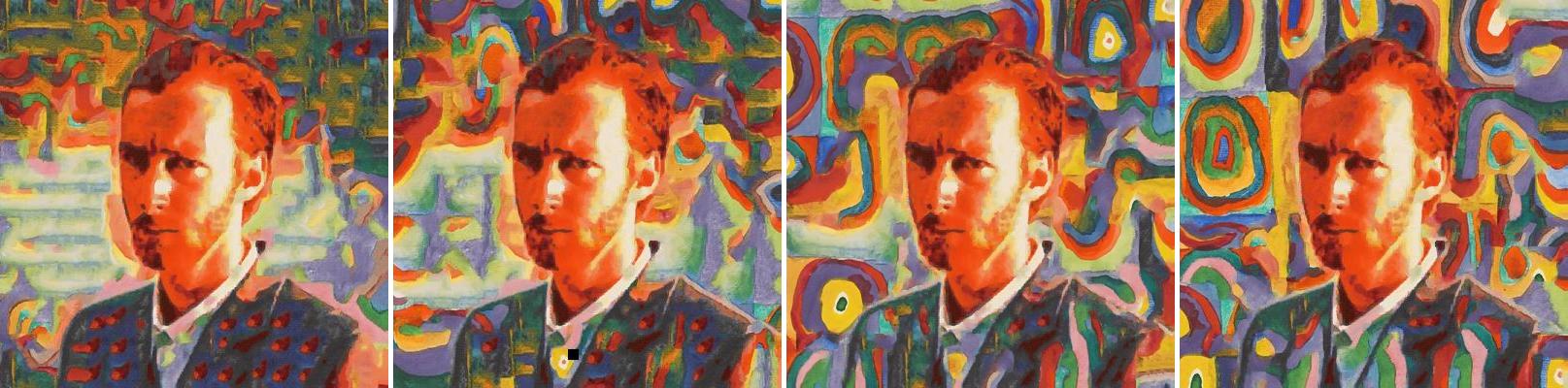}{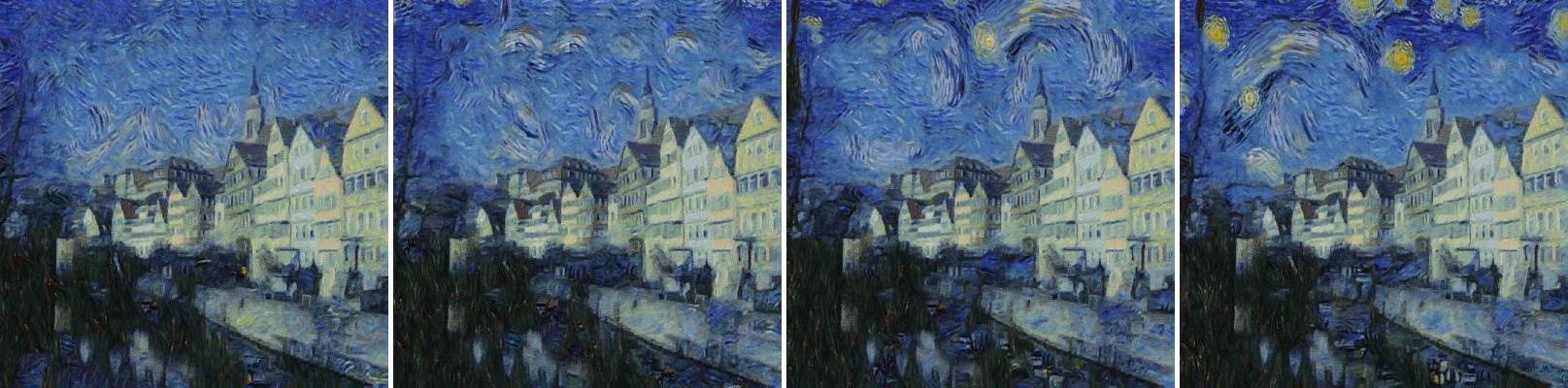}{1.6}{6.4}{{\bf Effect of $L$ - the pyramid depth:} These results are obtained by running the proposed algorithm for varying $L$ from $4$ (left) to $1$ (right) on two examples. The original content and style images referring to these experiments are found in earlier reported results.}

Figure \ref{fig:neffect} presents two sets of results obtained by the proposed algorithm, where the difference is in the set of patch-sizes used. The full list of sizes is $n=[33,21,13,9,5]$, and the corresponding sub-sampling skips are $d=[28,18,8,5,3]$. The experiments performed are
\begin{enumerate}
\item experiment1 - $n=[33,21,13,9,5]$,
\item experiment2 - $n=[33,21,13,9]$,
\item experiment3 - $n=[33,21,13]$, and
\item experiment4 - $n=[33,21]$.
\end{enumerate}
\noindent We see that the most important difference appears in the content region, in which a more textured result is obtained when avoiding the smaller patch-sizes. Interestingly, we could suggest to apply the finer patch-sizes without the content-enforcement stage, thereby getting refinement of the final outcome (everywhere) without losing the textured foreground.

\myfigJPGTwo{neffect}{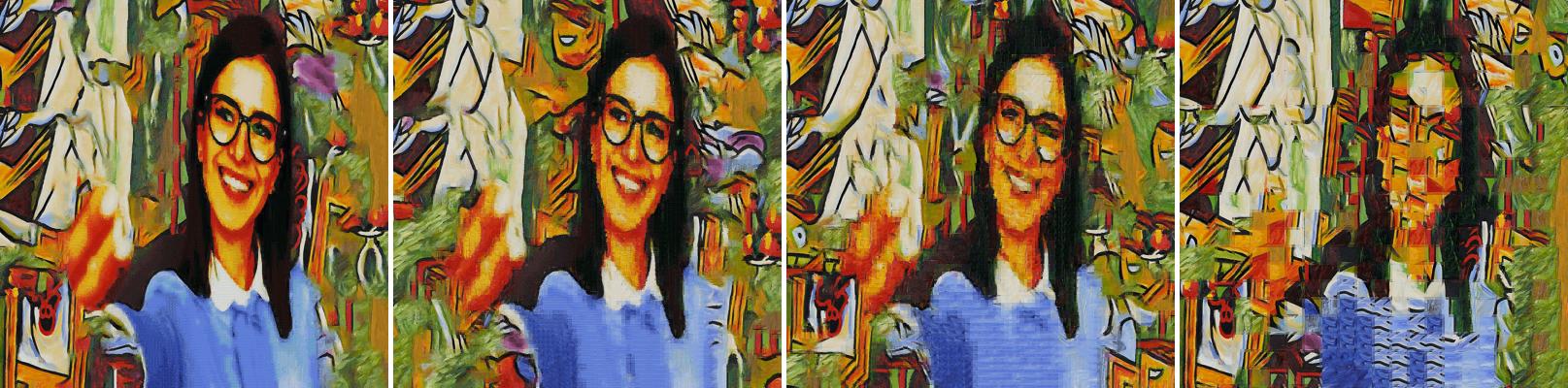}{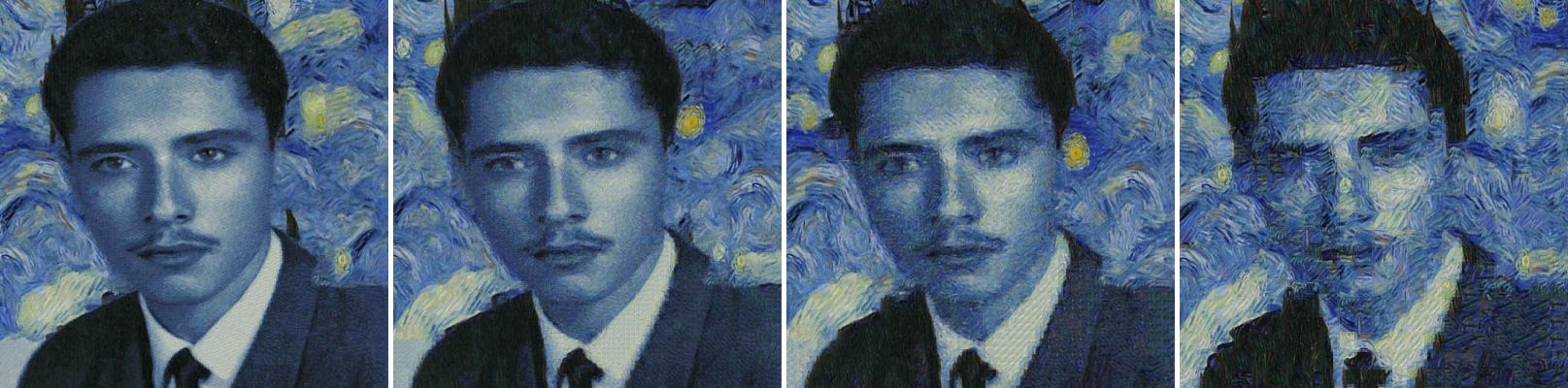}{1.6}{6.4}{{\bf Effect of $n$ - the patch-sizes used:} These results are obtained by running the proposed algorithm for varying set of patch-sizes from $n=[33,21,13,9,5]$ (left) to $n=[33,21]$ (right) on two examples. The original content and style images referring to these experiments are found in earlier reported results.}

Finally, we come to test the effect of the amount of overlap between adjacent patches in the sub-sampled grid $\Omega$. We apply the proposed algorithm with the patches-sizes $n=[33,21,13,9]$, and change $d$ to be more and more dense, by the following settings:
\begin{enumerate}
\item experiment1 - $d=[28,18,11,8]$ ($\approx \frac{n}{1.2}$),
\item experiment2 - $d=[23,15,9,6]$ ($\approx \frac{n}{1.2^2}$),
\item experiment3 - $d=[16,10,6,4]$ ($\approx \frac{n}{1.2^4}$), and
\item experiment4 - $d=[8,5,3,2]$ ($\approx \frac{n}{1.2^8}$).
\end{enumerate}
\noindent The conclusion from this experiment is very clear -- there is almost no visual benefit in using a denser grid $\Omega$, as the results obtained remain almost the same for small amount of overlap\footnote{The changes that are seen are due to the randomness of the algorithm.}. Naturally, since the density of $\Omega$ has an immediate impact on the run-time (the complexity of the algorithm is linear with the density), we prefer to use coarse sampling, as indeed practiced in all our reported experiments.

\myfigJPG{deffect}{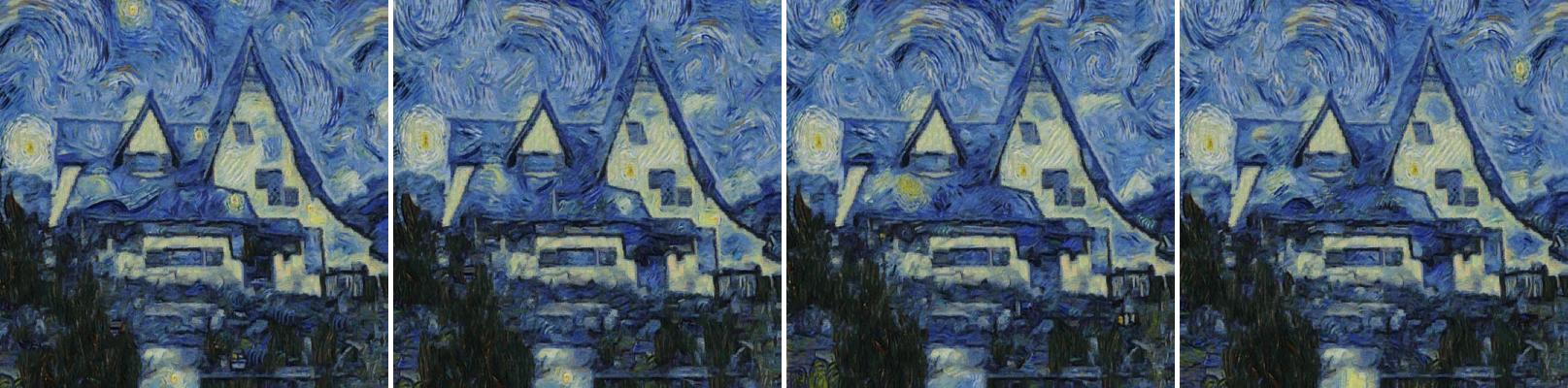}{1.6}{6.4}{{\bf Effect of $d$ - the subsampling:} These results are obtained by running the proposed algorithm for varying density of subsampling from very coarse (left) to relatively dense (right). The original content and style images referring to these experiments are found in earlier reported results.}

\subsection{Computational Complexity}

The proposed algorithm has been implemented in Matlab (2015b) without exploiting parallel options or Mex accelerations. The reported tests were performed on a Desktop Windows machine (3.5GHz E5-1650 v3 Intel Xeon CPU, 32 GByte RAM, 64-bits operating system). In measuring the running time of this algorithm, we assume that all the tree files for the fast-nearest-neighbor are pre-loaded to memory (this takes $20-30$ seconds). The run-time for the whole implementation stands on $25-50$ seconds for an image of size $400 \times 400$ pixels. The varying times are due to possible differences in the patch-sizes used. Indeed, the most demanding part of the algorithm is the patch-matching, and especially the one that takes place in the very last full-resolution iterations.

When compared to Gatys' algorithm \cite{CNN1}, the proposed scheme is much faster and simpler, and being of roughly the same complexity as the algorithm proposed by Frigo et. al. \cite{Frigo}. When compared to the more recent CNN methods that train a feed-forward network to perform the style-transfer \cite{CNN2,CNN3,CNNN4}, it is hard to make claims about run-times as the implementations are very different. However, a clear benefit of our algorithm is the ability to work with any pair without pre-training, as the convolutional neural networks require.

%%%%%%%%%%%%%%%%%%%%%%%%%%%%%%%%%%%%%%%%%%%%%%%%%%%%%%%%%%%%%%%%%%%%%%%%%%%%%%%%
%%%%%%%%%%%%%%%%%%%%%%%%%%%%%%%%%%%%%%%%%%%%%%%%%%%%%%%%%%%%%%%%%%%%%%%%%%%%%%%%

\section{Conclusions}\label{sec:Conclusions}

This paper puts forward a novel texture-synthesis-based solution to the style-transfer problem. The foundations of the proposed algorithm are adopted from the texture-synthesis algorithm by Kwatra et. al. \cite{Kwatra2005}, and those are augmented by proper modifications/additions in order to achieve the new goal of transferring style from one image to the content of another.
This work shows that effective and satisfactory style-transfer is within reach with this paradigm.

A fundamental question that accompanied us during this work and the extensive experiments is {\em When can we expect the algorithm to succeed?}, or {\em What constitutes a successful outcome?} These questions brought us to identify various additional avenues for further improvements to this algorithm. Among these, we would like to highlight the following few:
\begin{itemize}
\item We learned of the good and the bad effects that the palette-transfer stage brings to the style-transfer task. Further work is required to better design this stage and tailor it to lead to better quality transfer results. Our treatment of this part so far has been shallow, essentially adopted an existing solution.
\item The match between the scale of objects appearing in the style and the content images was also found to play a central role in the tendency of the transfer process to lead to pleasing result. That being said, all the tests reported here assumed fixed size images of $400 \times 400$ pixels, both for the style and the content, and this means that other sizes are scaled to this size. The transfer results are expected to be entirely different as we change the scale of one image versus the other, and more study is required here for better understanding this effect, and how to use it in order to choose automatically the scale of one versus the other that will lead to a successful outcome.
\item Our algorithm relies on segmentation, although we have shown that even without this stage, it may operate well under some circumstances. This calls for more work in order to remove the need for segmentation altogether, while preserving the quality of the results obtained.
\item Another delicate matter is the possibility that this algorithm (and indeed, any other texture-synthesis based method) would shamelessly copy pieces from the style image. The way to prevent this is to enrich the patch data-set, and this could be done in various ways, some of them simple (rotations, scaled versions, etc.), and some are more sophisticated (e.g. running a CNN-based texture-synthesis on the style image via \cite{CNN0} in order to get many more feasible patches).
\item Last but not least, our algorithm is still quite conservative when handling segmented content parts of the image. We envision modifications to the algorithm that will allow changing the content while preserving its essence (clearly, this calls for a clear definition of what is important and what is not in a given content).
\end{itemize}
\noindent We enjoyed working on this project, perhaps due to the artistic nature of the quest. We hope that the readers of this paper will shape this joy.

%%%%%%%%%%%%%%%%%%%%%%%%%%%%%%%%%%%%%%%%%%%%%%%%%%%%%%%%%%%%%%%%%%%%%%%%%%%%%%%%
\appendix
\section{Auxiliary Results}

As a followup to Figure 2 in the Introduction, which shows how the algorithm operates when there is no content image, here are few more such results.

\myfigJPGThree{NoContent2}{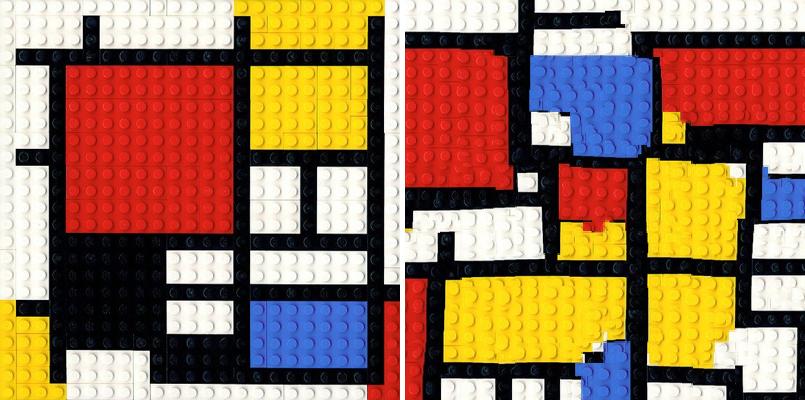}{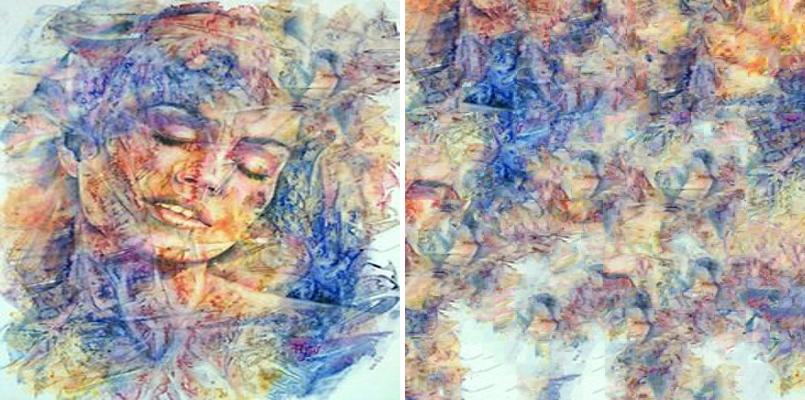}
{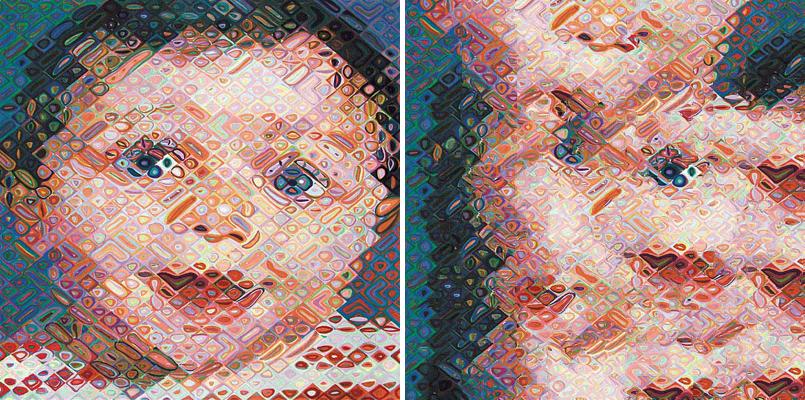}{2}{4}{{\bf No-Content Style-Transfer:} Style image (left), and its hallucination result (right) by Style-transfer applied with an empty content image.}

%%%%%%%%%%

We move now to the results section. As a followup to Figure 4, which bring face-segmented content images and their style-transfer results, we bring more results here, organized in three figures - \ref{fig:FaceSegment2}, \ref{fig:FaceSegment3} and  \ref{fig:FaceSegment4}.

\myfigJPGFour{FaceSegment2}{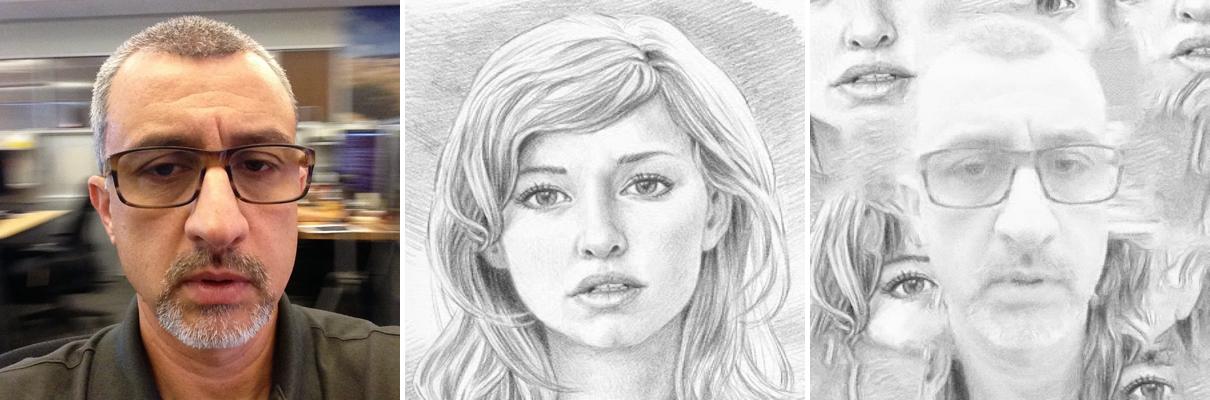}{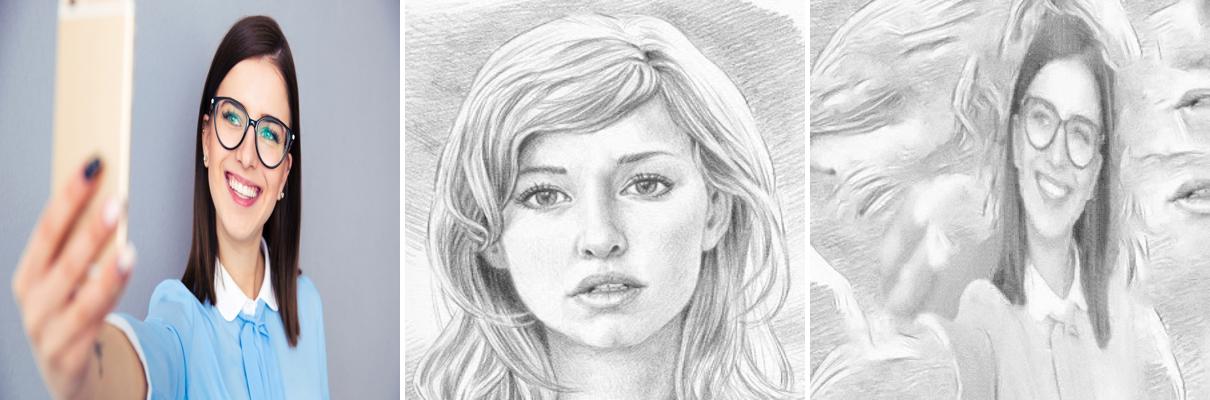}{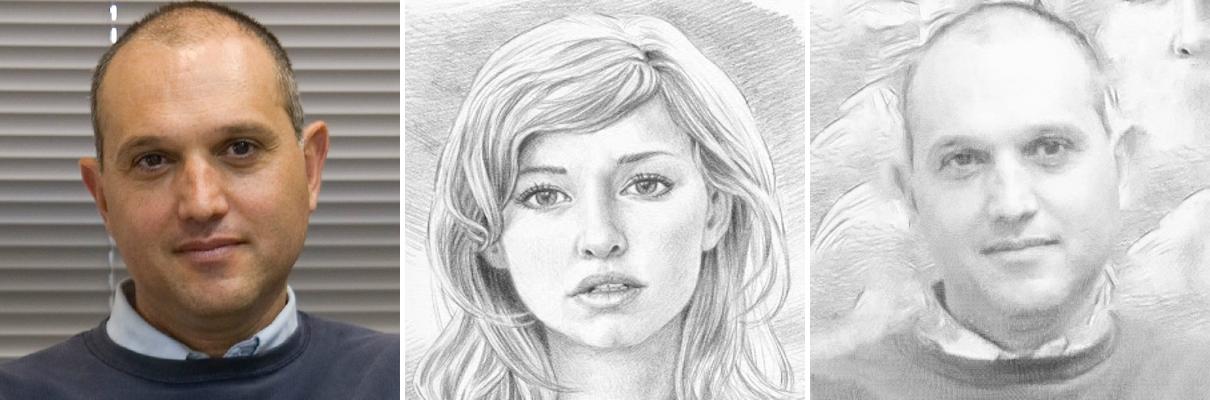}
{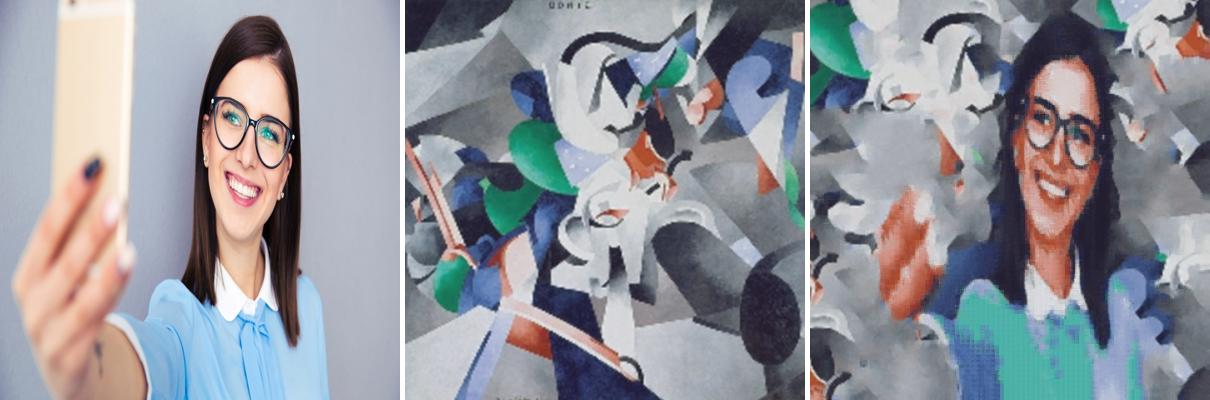}{1.6}{4.8}{{\bf Style-Transfer via Face-Segmentation:} Content image (left), Style image (middle), and the style-transfer result obtained (right). All these images contain faces, and the segmentation applied was Face-Based.}

\myfigJPGFour{FaceSegment3}{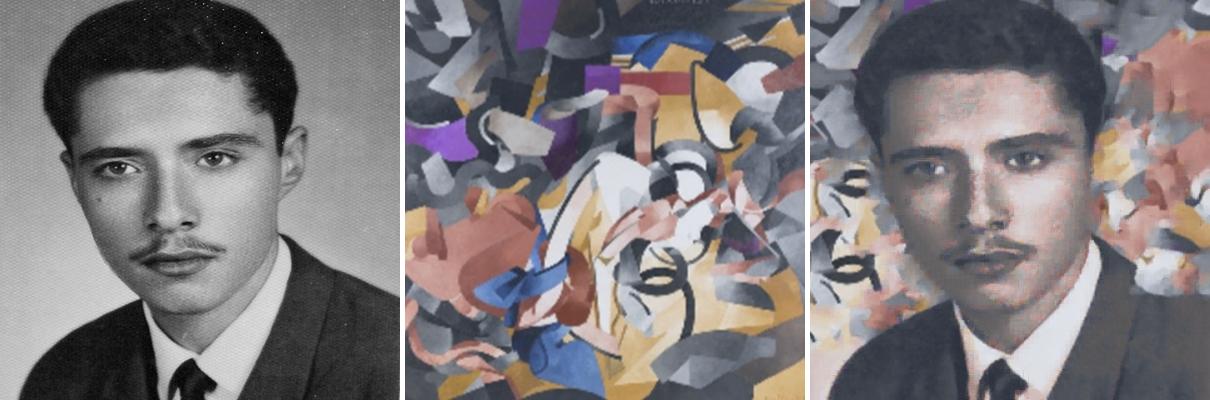}{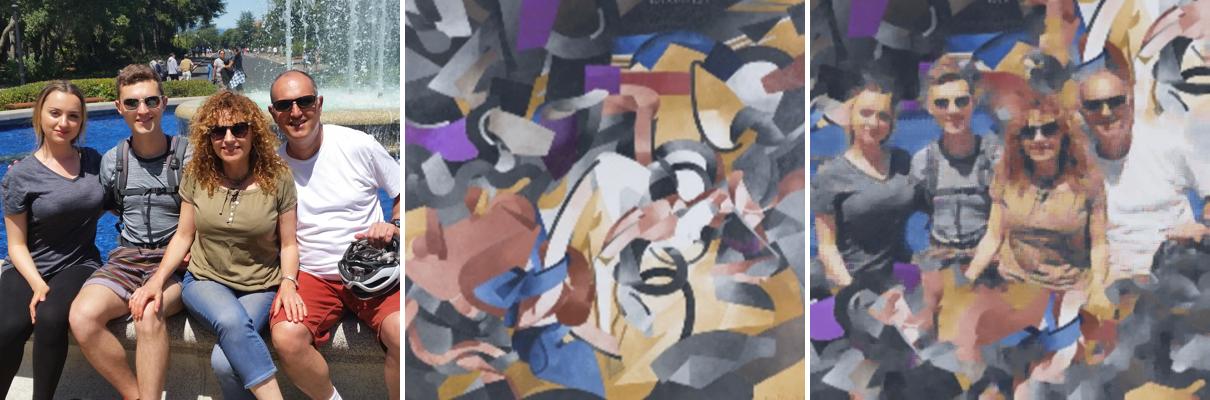}{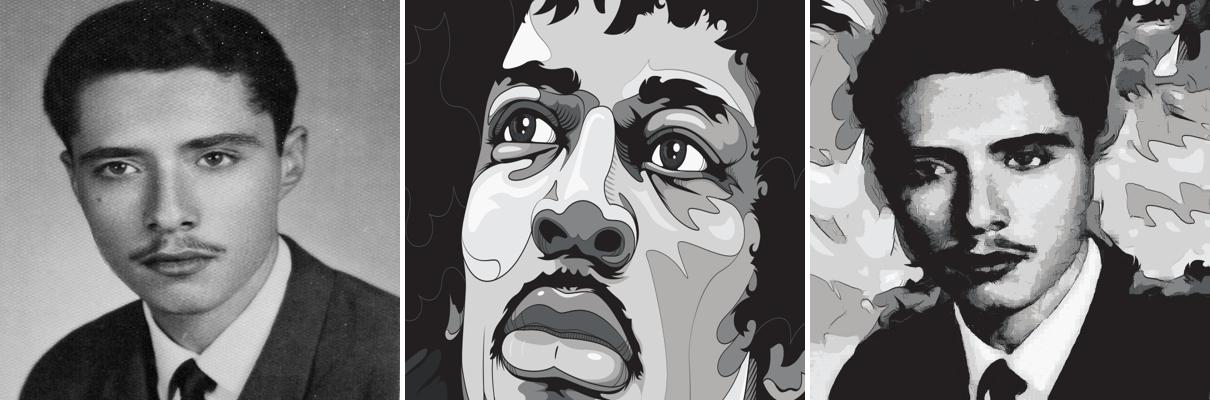}
{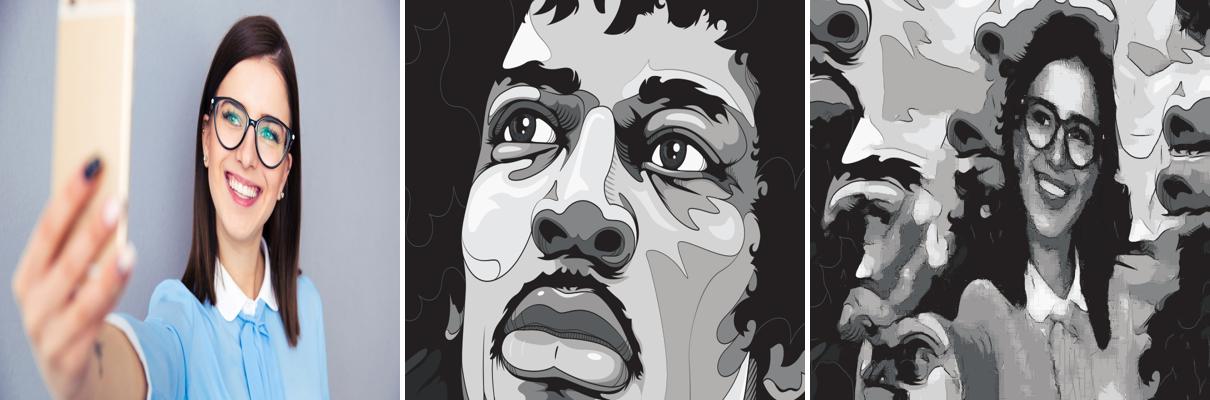}{1.6}{4.8}{{\bf Style-Transfer via Face-Segmentation:} Content image (left), Style image (middle), and the style-transfer result obtained (right). All these images contain faces, and the segmentation applied was Face-Based.}

\myfigJPGFour{FaceSegment4}{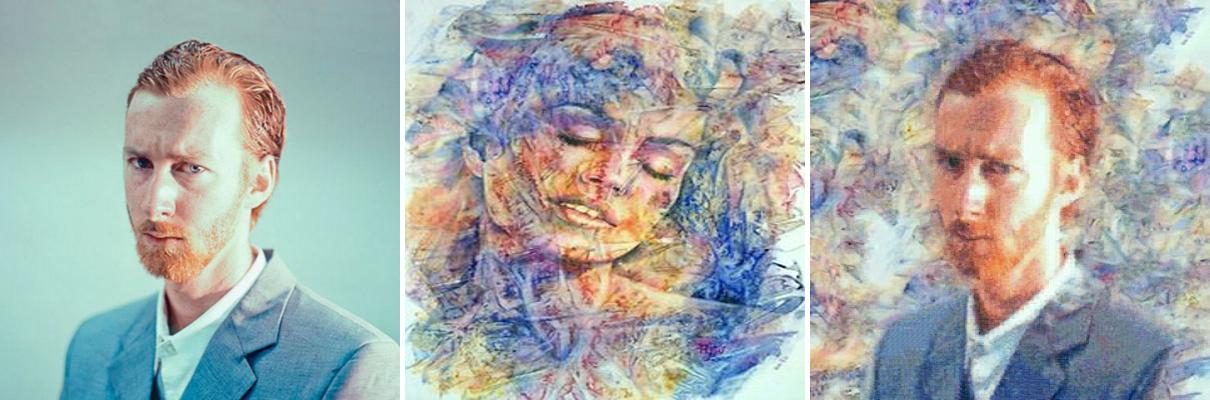}{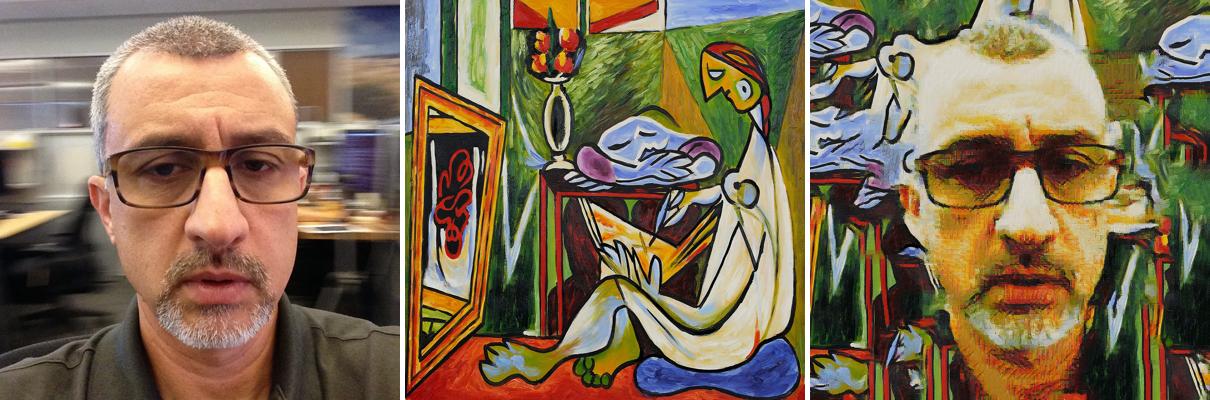}{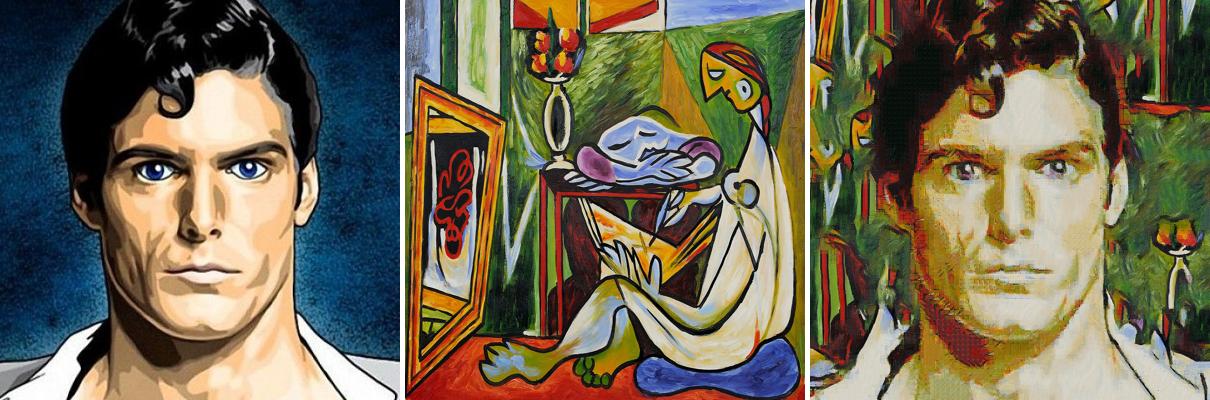}
{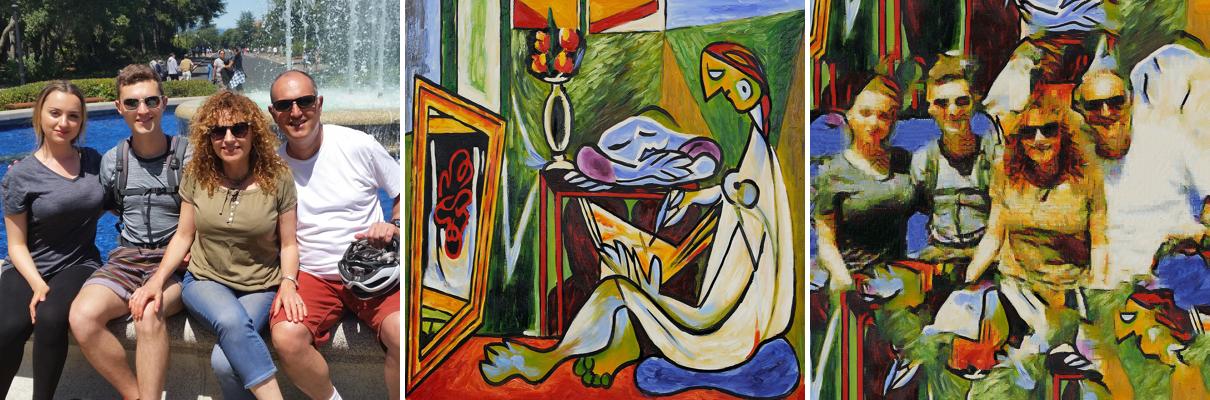}{1.6}{4.8}{{\bf Style-Transfer via Face-Segmentation:} Content image (left), Style image (middle), and the style-transfer result obtained (right). All these images contain faces, and the segmentation applied was Face-Based.}

\myfigJPGFour{FaceSegment5}{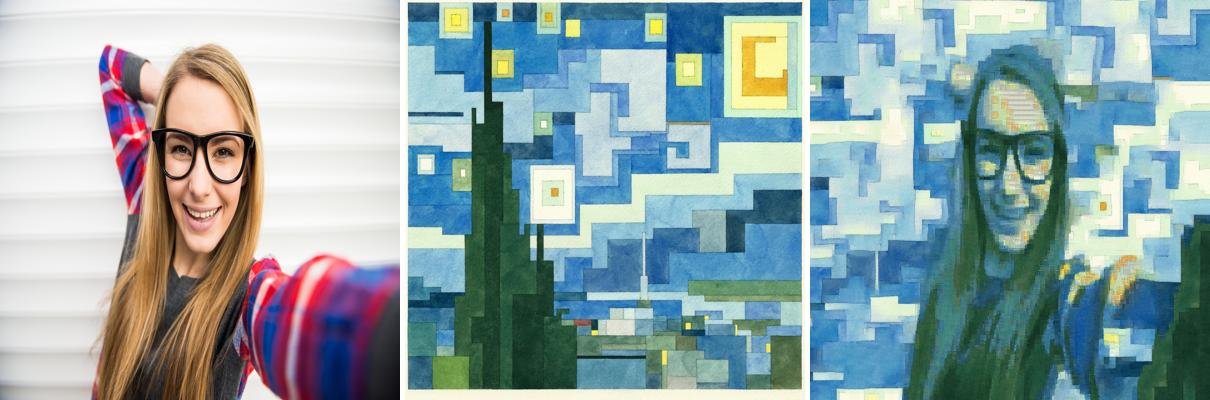}{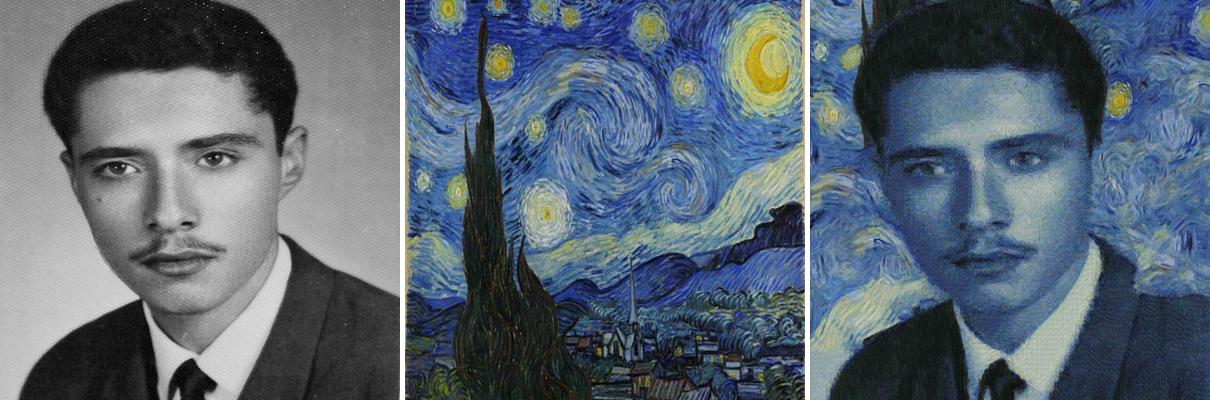}{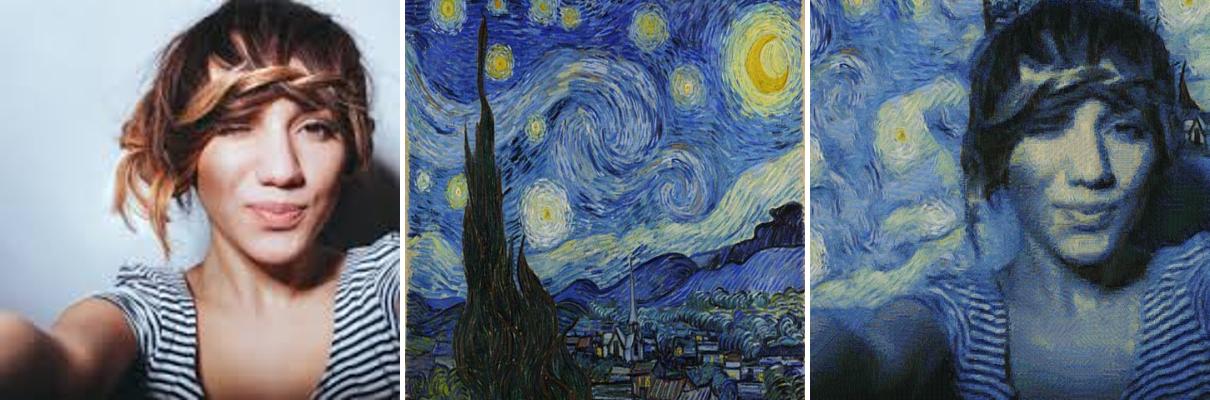}{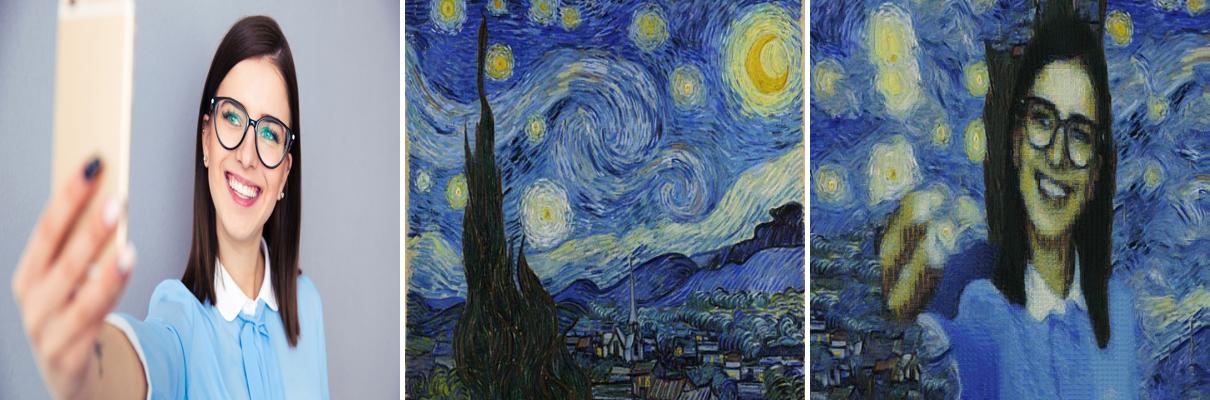}
{1.6}{4.8}{{\bf Style-Transfer via Face-Segmentation:} Content image (left), Style image (middle), and the style-transfer result obtained (right). All these images contain faces, and the segmentation applied was Face-Based.}

%%%%%%%%%%%

As a followup to Figure 5, which bring Edge-Based-segmented content images and their style-transfer results, we bring more results here, organized in two figures - \ref{fig:EdgeSegment2} and \ref{fig:EdgeSegment3}.

\myfigJPGFive{EdgeSegment2}{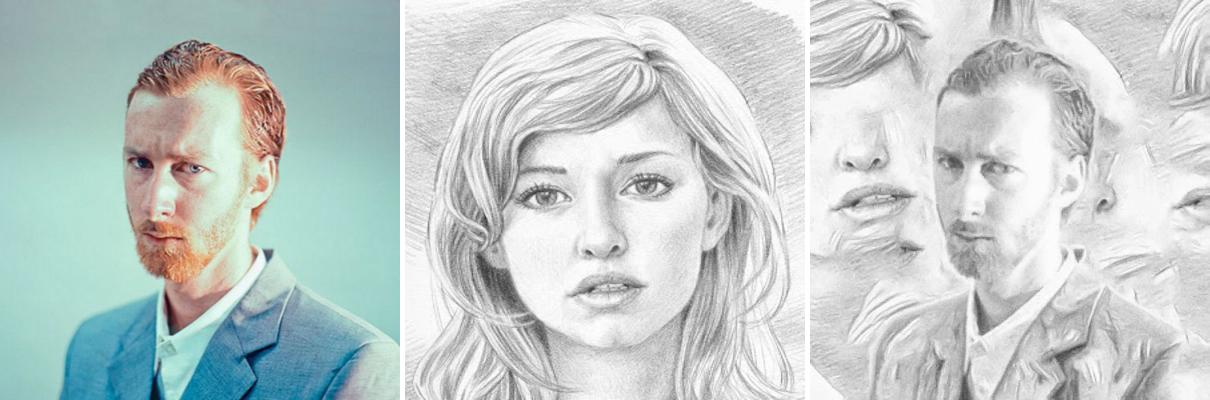}{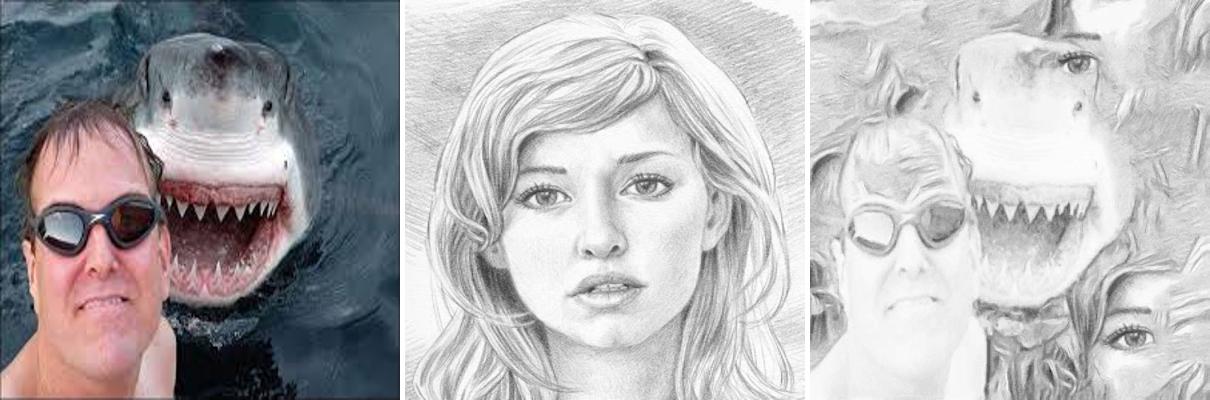}{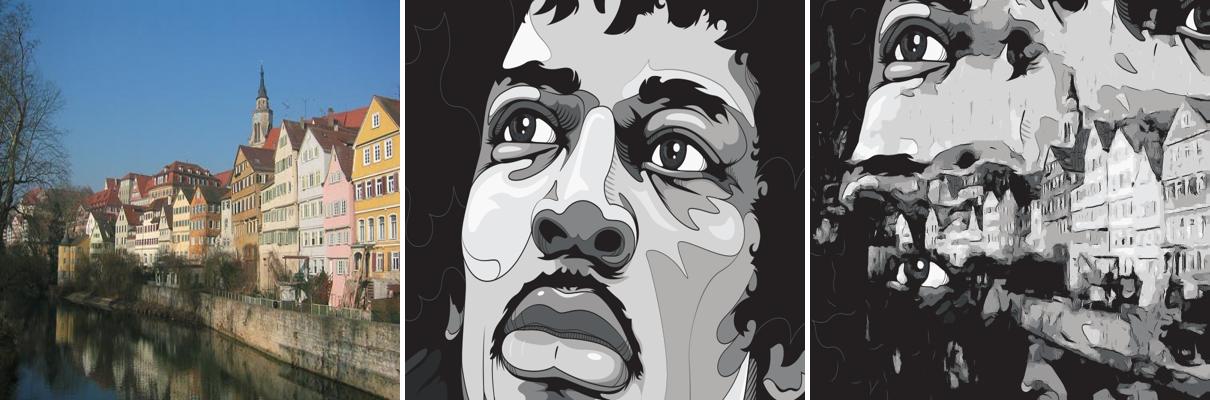}{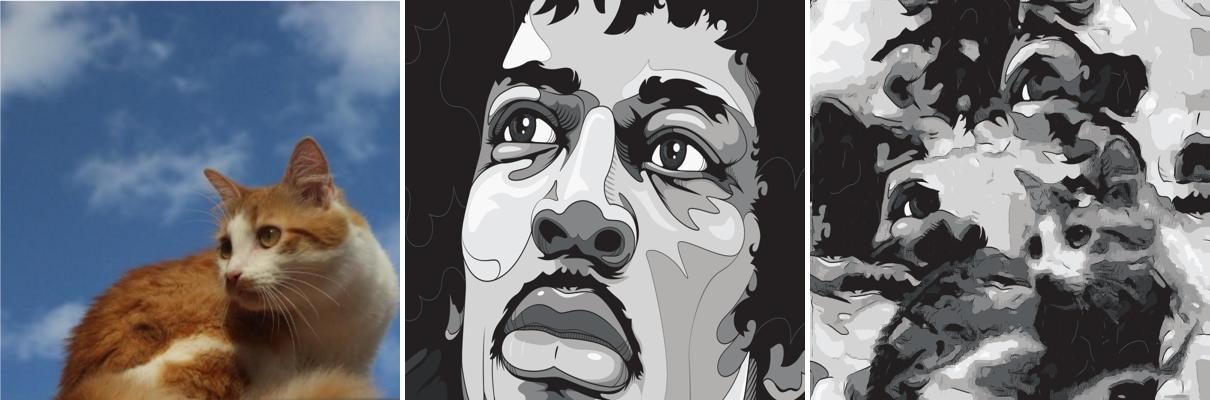}
{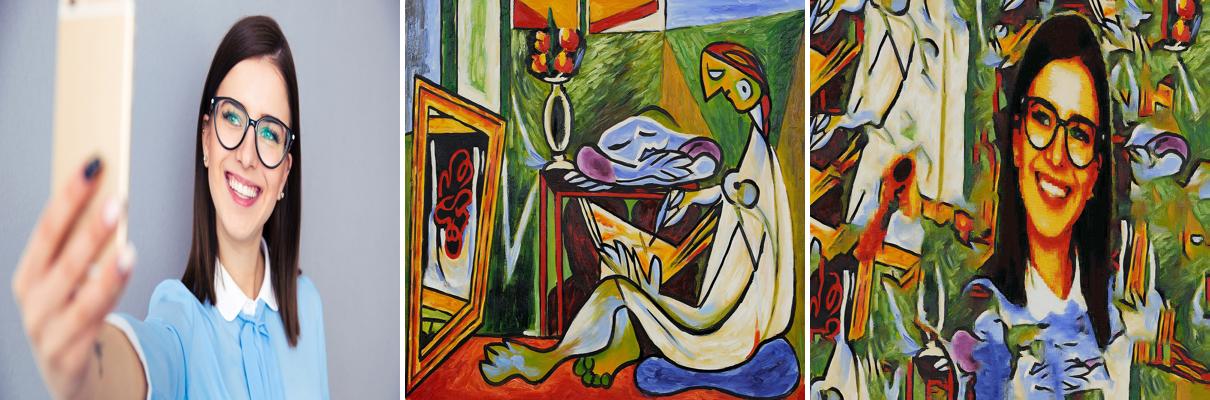}{1.6}{4.8}{{\bf Style-Transfer via Edge-Segmentation:} Content image (left), Style image (middle), and the style-transfer result obtained (right). All these images were segmented using the Edge-Based method.}

\myfigJPGFive{EdgeSegment3}{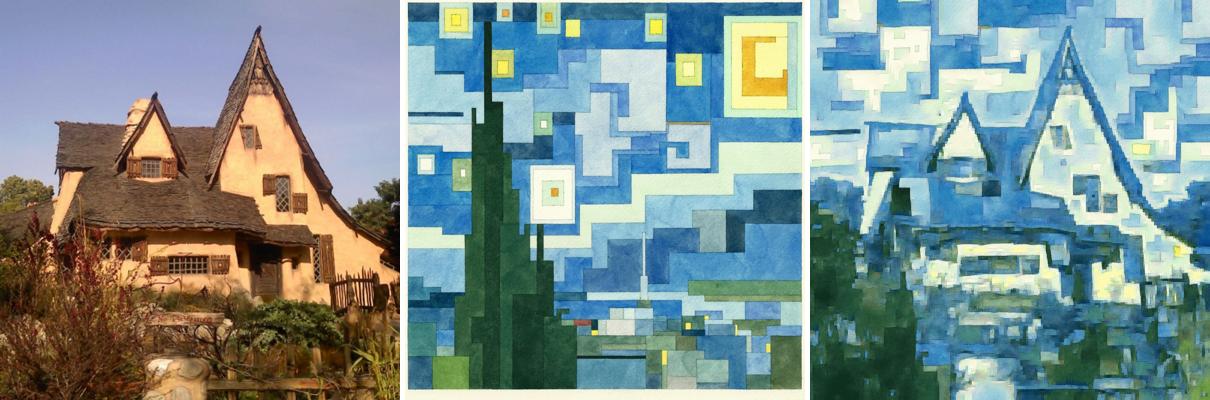}{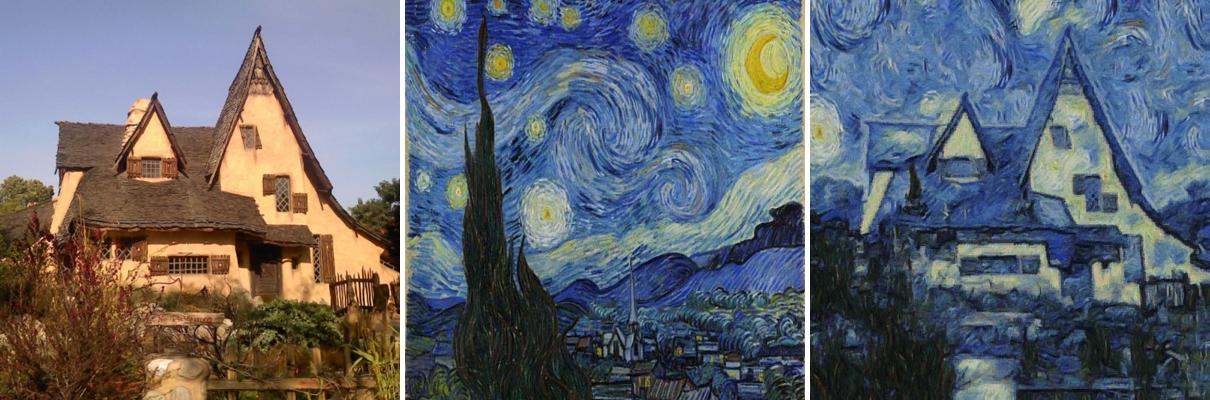}{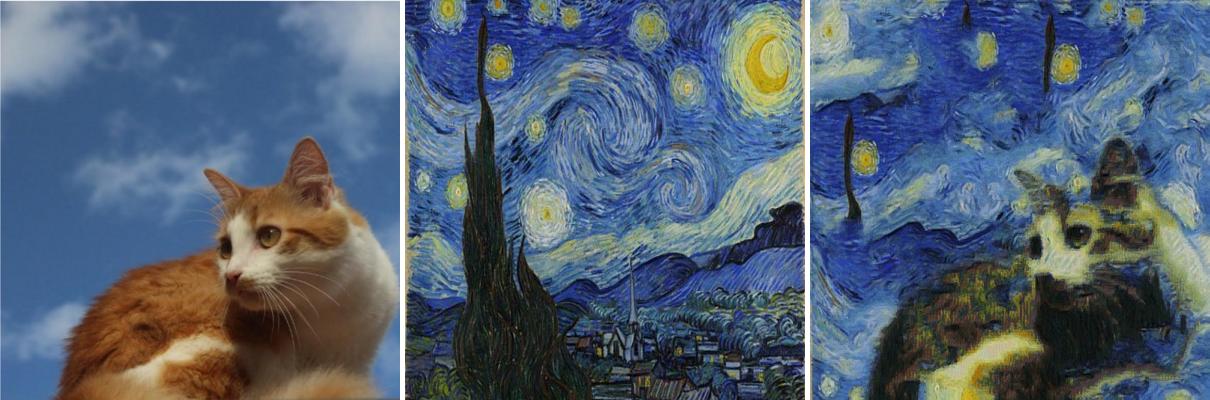}{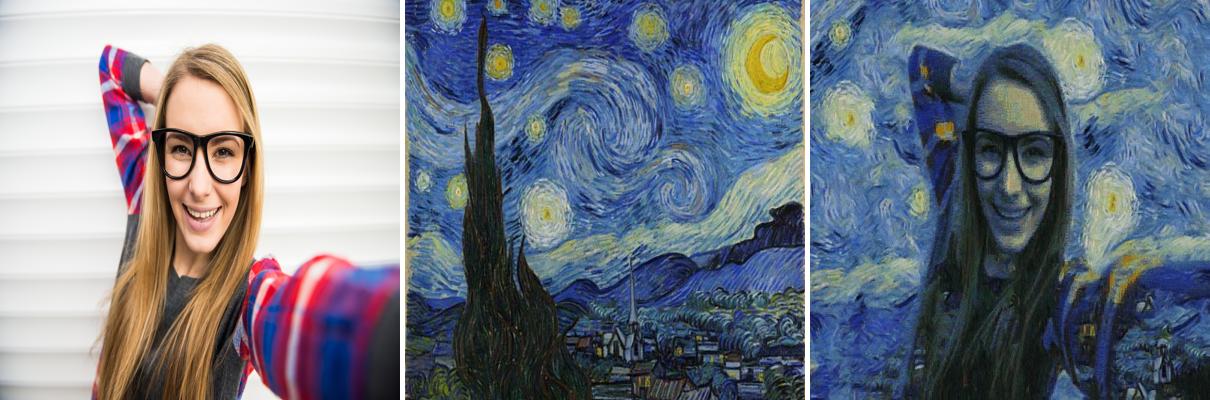}
{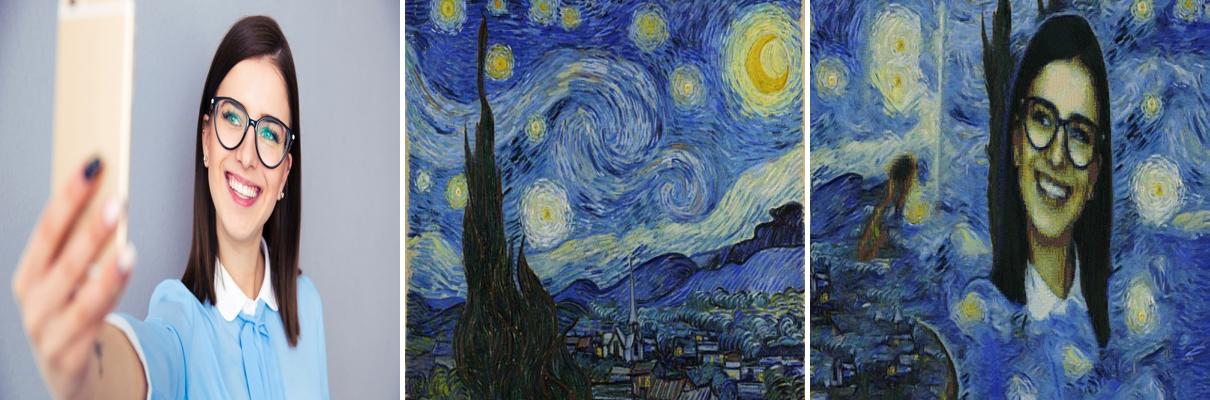}{1.6}{4.8}{{\bf Style-Transfer via Edge-Segmentation:} Content image (left), Style image (middle), and the style-transfer result obtained (right). All these images were segmented using the Edge-Based method.}

%%%%%%%%%%%

As a followup to Figure 6, which bring non-segmented content images and their style-transfer results, we bring more results here, in Figure \ref{fig:NoSegment2}.

\myfigJPGTwo{NoSegment2}{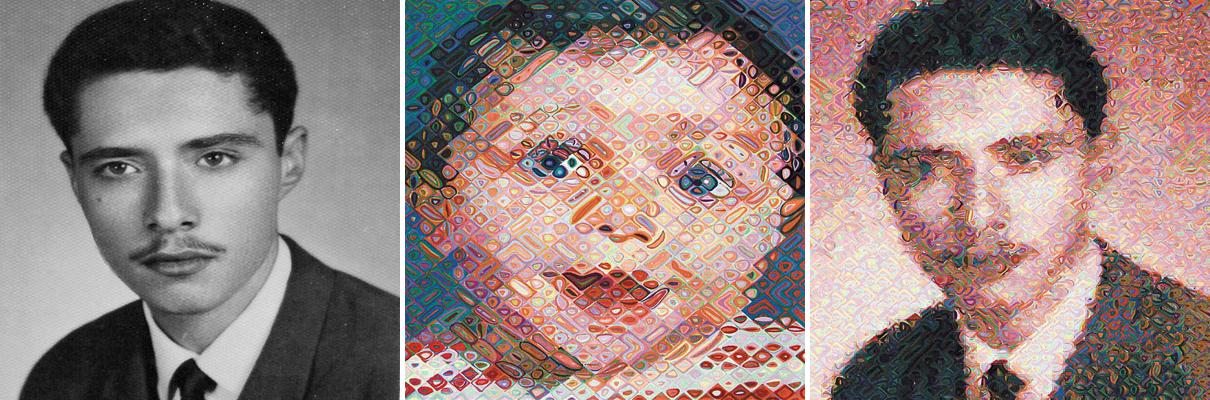}{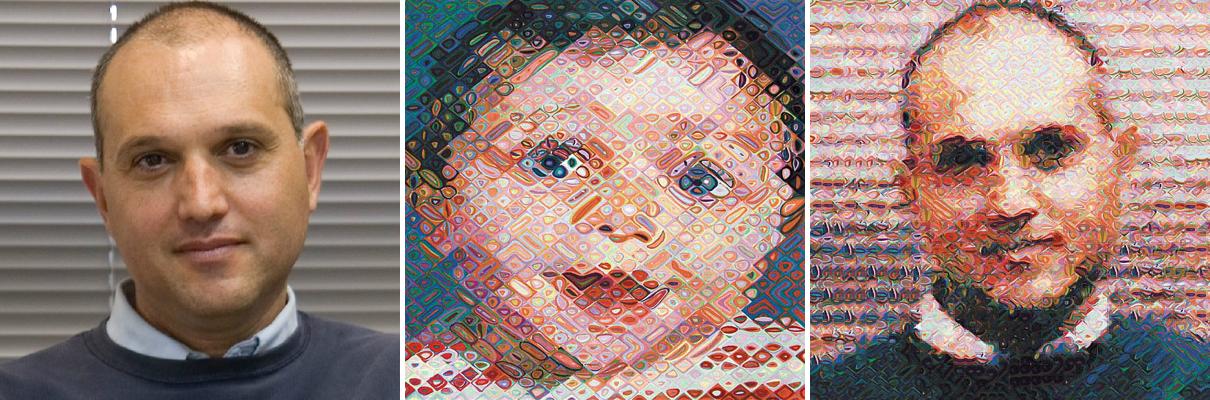}{1.6}{4.8}{{\bf Style-Transfer with No-Segmentation:} Content image (left), Style image (middle), and the style-transfer result obtained (right). All these images were segmented using no segmentation.}

%%%%%%%%%%%

As a followup to Figure 7, which shows the tendency of our algorithm to result with different outcomes when ran several times due to its random nature, we bring more results here, in Figure \ref{fig:Randomness2}.

\myfigJPG{Randomness2}{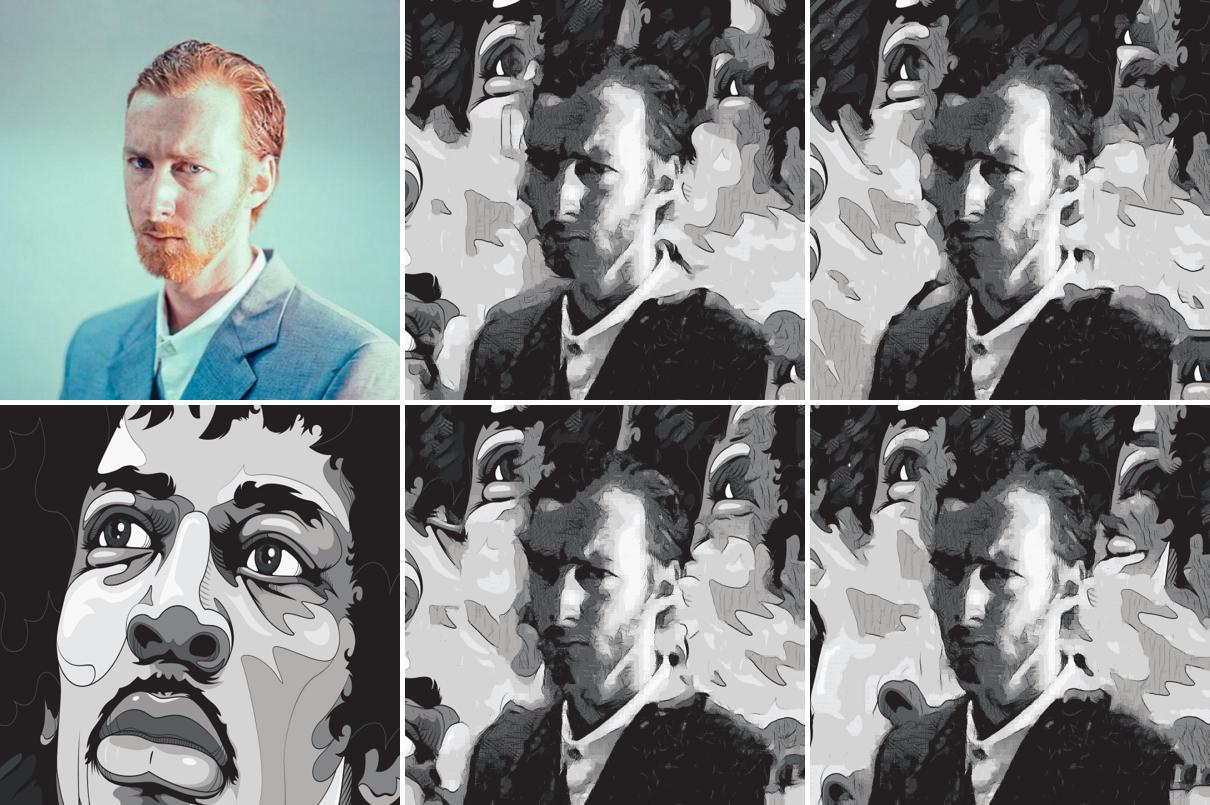}{3.2}{4.8}{{\bf Tendency to Randomness:} Content image (top-left), Style image (bottom-left), and the style-transfer results obtained by running the algorithm four times using the very same settings, and leaning on Face-Based segmentation.}

\bibliographystyle{ieee}

\end{document}